\documentclass[letterpaper]{article} 
\usepackage{aaai23}  
\usepackage{times}  
\usepackage{helvet}  
\usepackage{courier}  
\usepackage[hyphens]{url}  
\usepackage{graphicx} 
\usepackage{natbib}  
\usepackage{caption} 
\usepackage{algorithmic}
\usepackage{amssymb,amsmath}
\usepackage{amsfonts}
\usepackage{amsthm}
\usepackage[ruled,vlined,linesnumbered]{algorithm2e}
\usepackage{subfigure}

\usepackage{newfloat}
\usepackage{listings}
\DeclareCaptionStyle{ruled}{labelfont=normalfont,labelsep=colon,strut=off} 
\title{Robust Sequence Networked Submodular Maximization}
\author{
    Qihao Shi\textsuperscript{\rm 1\rm 2}, Bingyang Fu\textsuperscript{\rm 1}, Can Wang\textsuperscript{\rm 1}\thanks{Corresponding Author.}, Jiawei Chen\textsuperscript{\rm 1}, Sheng Zhou\textsuperscript{\rm 1}, Yan Feng\textsuperscript{\rm 1}, Chun Chen\textsuperscript{\rm 1}
}
\affiliations{
    \textsuperscript{\rm 1}School of Computing and Computer Science, Zhejiang University, Hangzhou, China\\
    \textsuperscript{\rm 2}School of Computing and Computer Science, Zhejiang University City College, Hangzhou, China \\
    \{shiqihao321, todobe, wcan, sleepyhunt, zhousheng\_zju, fengyan, chenc\}@zju.edu.cn
}

\usepackage{bibentry}

\begin{document}

\maketitle

\begin{abstract}
In this paper, we study the \underline{R}obust \underline{o}ptimization for \underline{se}quence \underline{Net}worked \underline{s}ubmodular maximization (RoseNets) problem. We interweave the robust optimization  with the sequence networked submodular maximization. The elements are connected by a directed acyclic graph and the objective function is not submodular on the elements but on the edges in the graph. Under such networked submodular scenario, the impact of removing an element from a sequence depends both on its position in the sequence and in the network. This makes the existing robust algorithms inapplicable. In this paper, we take the first step to study the RoseNets problem. We design a robust greedy algorithm, which is robust against the removal of an arbitrary subset of the selected elements. The approximation ratio of the algorithm depends both on the number of the removed elements and the network topology. We further conduct experiments on real applications of recommendation and link prediction. The experimental results demonstrate the effectiveness of the proposed algorithm.
\end{abstract}

\newtheorem{lemma}{\textbf{Lemma}}
\newtheorem{definition}{\textbf{Definition}}
\newtheorem{theorem}{\textbf{Theorem}}
\newtheorem{problem}{\textbf{Problem}}
\renewcommand{\proofname}{\rm \textbf{Proof}}

\section{Introduction}
Submodularity is an important property that models a diminishing return phenomenon, i.e., the marginal value of adding an element to a set decreases as the set expands. It has been extensively studied in the literature, mainly accompanied with maximization or minimization problems of set functions \cite{nemhauser1978best,khuller1999budgeted}. This is called submodularity. Mathematically, a set function $f: 2^V \to \mathbb{R}$ is submodular if for any two sets $A\subseteq B \subseteq V$ and an element $v\in V\backslash B$, we have $f(A\cup v) - f(A) \ge f(B\cup v) -f(B)$. Such property finds a wide range of applications in machine learning, combinatorial optimization, economics, and so on. \cite{krause2005near,lin2011class,Li2022HERO,shi2021profit,kirchhoff2014submodularity,gabillon2013adaptive,kempe2003maximizing,wang2021efficient}.

Further equipped with monotonicity, i.e., $f(A)\le f(B)$ for any $A\subseteq B\subseteq V$, a submodular set function can be maximized by the cardinality constrained classic greedy algorithm, achieving an approximation ratio up to $1-1/e$ \cite{nemhauser1978best} (almost the best). Since then, the study of submodular functions has been extended by a variety of different scenarios, such as non-monotone scenario, adaptive scenario, and continuous scenario, etc \cite{feige2007maximizing,golovin2011adaptive,das2011submodular,bach2019submodular,shi2019adaptive}.

The above works focus on the set functions. In real applications, the order of adding elements plays an important role and affects the function value significantly. Recently, the submodularity has been generalized to sequence functions \cite{zhang2015string,tschiatschek2017selecting,streeter2008online,zhang2013near}. Considering sequences instead of sets causes an exponential increase in the size of the search space, while allowing for much more expressive models.

In this paper, we consider that the elements are networked by a directed graph. The edges encode the additional value when the connected elements are selected in a particular order. Such setting is not given a specific name before. To distinguish from the classic submodularity, we in this paper name it as networked submodularity (Net-submodularity for short). More specifically, the Net-submodular function $f(\sigma)$ is a sequence function, which is not submodular on the induced element set by $\sigma$ but is submodular on the induced edge set by $\sigma$. The Net-submodularity is first considered in \cite{tschiatschek2017selecting}, which mainly focuses on the case where the underlying graph is a directed acyclic graph. General graphs and hypergraphs are considered in \cite{mitrovic2018submodularity}.

Recently, robust versions of the submodular maximization problem have arisen \cite{orlin2018robust,mitrovic2017streaming,bogunovic2017robust,sallam2020robust} to meet the increasing demand in the stability of the system. The robustness of the model mainly concerns with its ability in handling the malfunctions or adversarial attacks, i.e., the removal of a subset of elements in the selected set or sequence. Sample cases of elements removal in real world scenarios include items sold out or stop production in recommendation \cite{mitrovic2018submodularity}, web failure of user logout in link prediction \cite{mitrovic2019adaptive} and equipment malfunction in sensor allocation or activation \cite{zhang2015string}. In this paper, we take one step further and study a new problem of \underline{ro}bust \underline{se}quence \underline{net}worked \underline{s}ubmodular maximization (RoseNets). We show an example in Figure 1 to illustrate the importance of RoseNets problem.

See in Figure 1. Suppose all edge weights in sequence A are 0.9, in sequence B are 0.5, in sequence C are 0.4. Let the net-submodular utility function $f$ of the sequence be the summation of all the weights of the induced edge set by the sequence. Such utility function is obviously monotone but not submodular.\footnote{Easy to see that in sequence B, the utility of $\{B4\}$ is 0, but $\{B3,B4\}$ is 0.5, which violates the submodularity.} However, it is submodular on the edge set. Now we can see, the utility of sequence A, B and C are 2.7 (largest), 2.5 and 2.4 respectively. We can easily check that if one node would be removed in each sequence, the worst utility after removal of sequence A, B and C is 0.9, 1.0 (largest), and 0.8. If we remove two nodes in each sequence, the utility of A, B and C becomes 0, 0, and 0.4 (largest). With different number of nodes removed, the three sequences show different robustness. Existing non-robust algorithm may select sequence A since it has the largest utility. However, sequence B and C are more robust against node removal.

\begin{figure}
  \centering
  \includegraphics[width=2.5in]{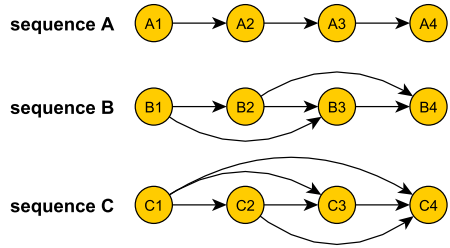}\\
  \caption{Example of RoseNets}
\vspace{-1.5em}
\end{figure}

Given a net-submodular function and the corresponding network, the RoseNets problem aims to select a sequence of elements with cardinality constraints, such that the value of the sequence function is maximized when a certain number of the selected elements may be removed. As far as sequence functions and net-submodularity are concerned, the design and analysis of robust algorithms are faced with novel technical difficulties. The impact of removing an element from a sequence depends both on its position in the sequence and in the network. This makes the existing robust algorithms inapplicable here. It is unclear what conditions are sufficient for designing efficient robust algorithm with provable approximation ratios for RoseNets problem. We aim to take a step for answering this question in this paper. Our contributions are summarized as follows.

\begin{enumerate}
\item To the best of our knowledge, this is the first work that considers the RoseNets problem. Combining robust optimization and sequence net-submodular maximization requires subtle yet critical theoretical efforts.
\item We design a robust greedy algorithm that is robust against the removal of an arbitrary subset of the selected sequence. The theoretical approximation ratio depends both on the number of the removed elements and the network topology.
\item We conduct experiments on real applications of recommendation and link prediction. The experimental results demonstrate the effectiveness and robustness of the proposed algorithm, against existing sequence submodular baselines. We hope that this work serves as an important first step towards the design and analysis of efficient algorithms for robust submodular optimization.
\end{enumerate}

\section{Related Works}
Submodular maximization has been extensively studied in the literature. Efficient approximation algorithms have been developed for maximizing a submodular set function in various settings \cite{nemhauser1978best,khuller1999budgeted,calinescu2011maximizing,chekuri2014submodular}. By considering the robustness requirement, recently, robust versions of submodular maximization have been extensively studied. These works aim at selecting a set of elements that is robust against the removal of a subset of elements. The first algorithm for the cardinality constrained robust submodular maximization problem is studied in \cite{orlin2018robust}. A constant factored approximation ratio is achieved. The selected $k$-sized set is robust against the removal of any $\tau$ elements of the selected set. The constant approximation ratio is valid as long as $\tau= O(\sqrt{k}))$. An improvement is made in \cite{bogunovic2017robust}, which provide an algorithm that guarantees the same constant approximation ratio but allows the removal of a larger number of elements (i.e.,$\tau= O(k))$. With a mild assumption, the algorithm proposed in \cite{mitrovic2017streaming} allows the removal of an arbitrary number of elements. The restriction on $\tau$ is relaxed in \cite{tzoumas2017resilient}, while the derived approximation ratio is parameterized $\tau$. This work is extended to a multi-stage setting in \cite{tzoumas2018resilient} and \cite{tzoumas2020robust}. The decision at each stage would takes into account the failures that happened in the previous stages. Other constrains that are combined with the robust optimization include fairness, privacy issues and so on \cite{mirzasoleiman2017deletion,kazemi2018scalable}.

The concept of sequence (or string) submodularity for sequence functions is a generalization of submodularity, which has been introduced recently in several studies \cite{zhang2015string,streeter2008online,zhang2013near}
The above works all consider element-based robust submodular maximization. Networked submodularity is considered in  \cite{tschiatschek2017selecting,mitrovic2018submodularity}, where the sequential relationship among elements is encoded by a directed acyclic graph. Following the networked submodularity setting, the work in \cite{mitrovic2019adaptive} introduces the idea of adaptive sequence submodular maximization, which aims to utilize the feedback obtained in previous iterations to improve the current decision. In this paper, we follow the networked submodularity setting, and study the RoseNets problem. It is unclear whether all of the above algorithms can be properly extended to our problem, as converting a set function to a sequence function and submodularity to networked submodularity could result in an arbitrarily bad performance. Establishing the approximation guarantees for RoseNets problem would require a more sophisticated analysis, which calls for more in-depth theoretical efforts.

\section{System Model and Problem Definition}
In this paper, we follow the networked submodular sequence setting \cite{tschiatschek2017selecting,mitrovic2018submodularity}. Let $V = \{v_1,v_2,...,v_n\}$ be the set of $n$ elements. A set of edges $E$ represents that there is additional utility in picking certain elements in a certain order. More specifically, an edge $e_{ij} = (v_i,v_j)$ represents that there is additional utility in selecting $v_j$ after $v_i$ has already been chosen. Self-loops (i.e., edges that begin and end at the same element) represents that there is individual utility in selecting an element.

Given a directed graph $G = (V,E)$, a non-negative monotone submodular set function $h: 2^E \rightarrow \mathbb{R}_{\ge 0}$, and a parameter $k$, the objective is to select a non-repeating sequence $\sigma$ of $k$ unique elements that maximizes the objective function:
\[
f(\sigma)=h(E(\sigma)),
\]
where $E(\sigma)$ contains all the edges $(v_i,v_j)\in E$ that $v_i$ is select before $v_j$ in $\sigma$.

We say $E(\sigma)$ is the set of edges induced by the sequence $\sigma$. It is important to note that the function $h$ is a submodular set function over the edges, not over the elements. Furthermore, the objective function $f$ is neither a set function, nor is it necessarily submodular on the elements. We call such a function $f(\sigma)$ as a \textit{networked submodular} function.

We define $f(\sigma - Z)$ to represent the residual value of the objective function after the removal of elements in set $Z$. In this paper, the \underline{ro}bust \underline{se}quence \underline{net}worked \underline{s}ubmodular maximization (RoseNets) problem is formally defined below.

\begin{definition}
Given a directed graph $G=(V,E)$, a networked submodular function $f(\cdot)$ and robustness parameter $\tau$, the RoseNets problem aims at finding a sequence $\sigma$ such that it is robust against the worst possible removal of $\tau$ nodes:
\[
\max_{\sigma:|\sigma|\le k} \min_{Z\in \sigma,|Z|\le \tau} f(\sigma - Z).
\]
\end{definition}
The robustness parameter $\tau$ represents the size of the subset $Z$ that is removed. After the removal, the objective value should remain as large as possible. For $\tau = 0$, the problem reduces to the classic sequence submodular maximization problem \cite{mitrovic2018submodularity}.

\section{Robust Algorithm and Theoretical Results}
Direct applying the Sequence Greedy algorithm \cite{mitrovic2018submodularity} to solve the RoseNets problem would return an arbitrary bad solution. We can construct a very simple example for an illustration. See in Figure 2. Let the edge weights of $(A,B),(B,C),(B,E),(B,F)$ be 0.9 and $(C,D),(C,G),(D,G)$ be 0.5. Let the net-submodular utility function $f$ of the selected sequence be the summation of all weights of the induced edge set by the sequence. Suppose we are to select a sequence with 5 elements. Using the Sequence Greedy algorithm, sequence $\langle A,B,C,E,F \rangle$ will be selected\footnote{Suppose elements are selected in the alphabetic order if the edge weights are equal. Similar examples can be easily constructed when elements are selected at random.} for maximizing the utility, i.e., $(0.9\cdot 4=3.6)$. However, if $\tau=2$, i.e., two elements would be removed, removing $B$ and any other one element (worst case) makes the utility become 0.

\begin{figure}
  \centering
  \includegraphics[width=2.2in]{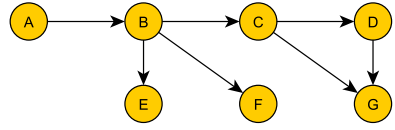}\\
  \caption{Example of Sequence Greedy}
\vspace{-1.5em}
\end{figure}

\subsection{RoseNets Algorithm}
We wish to design an algorithm that is robust against the removal of an arbitrary subset of $\tau$ selected elements. In this paper, we propose the RoseNets Algorithm, which can approximately solves the RoseNets problem and is shown in Algorithm 1. Note we consider the case that $k\ge 3$.

The limitation of the Sequence Greedy algorithm is that the selected sequence is vulnerable. The overall utility might be concentrated in the first few elements. Algorithm 1 is motivated by this key observation and works in two steps. In Step 1 (the first \textit{while} loop), we select a sequence $\sigma_1$ of $\tau$ elements from $V$ in a greedy manner as in Sequence Greedy. In Step 2 (the second \textit{while} loop), we select another sequence $\sigma_2$ of $k-\tau$ elements from $V\backslash \sigma_1$, again in a greedy manner as in Sequence Greedy. Note that when we select sequence $\sigma_2$, we perform the greedy selection as if sequence $\sigma_1$ does not exist at all. This ensures that the value of the final returned sequence $\sigma = \sigma_1 \oplus \sigma_2$ is not concentrated in either $\sigma_1$ or $\sigma_2$. The complexity of Algorithm 1 is $O(k|E|)$, which is in terms of the number of function evaluations used in the algorithm.

To show the differences and benefits of the RoseNets algorithm, we go back to see the example in Figure 2. When $k=5$ and $\tau=2$, the RoseNets algorithm will select the sequence $\sigma_1=\langle A,B \rangle$ and sequence $\sigma_2=\langle C,D,G \rangle$. When selecting $\sigma_2$, the RoseNets algorithm would not consider element $B$ in $\sigma_1$. Thus element $E$ and $F$ are regarded as making not contribution to the utility function. The RoseNets algorithm will return the sequence $\sigma=\langle A,B,C,D,G \rangle$. The worst case of removing $\tau=2$ elements, is to remove $B$ and any one element in $\{C,D,G\}$. The residual utility is 0.5. Remember the case for Sequence Greedy algorithm, the residual utility of the worst case is 0. This example shows the benefits of the RoseNets algorithm.

Both the examples in Figure 1 and Figure 2 imply that a robust sequence should have complex network structure. The utility should not concentrate in a center element, but aggregated from all the edges among elements in the sequence. Such a robust sequence can only be selected by multiple trials of greedy selection, with the trials neglecting each other. Otherwise, the central element (if exists) with its high edge weight neighbors are probability selected, as node $B$ in Figure 2. If such a node is removed, the utility would slump. In this paper, we implement such intuitive strategy by using two trials of greedy selection. Intuitively, invoking more times of greedy selection trial may improve the approximation ratio. However, as we are to take a first step in the RoseNets problem while our theoretical analysis is already non-trivial, we leave the design of multi-part selection algorithm and the approximation ratio analysis for future works.

\begin{algorithm}[t]
\small
\caption{RoseNets Algorithm}
$\sigma=\emptyset$, $\sigma_1=\emptyset$, $\sigma_2=\emptyset$\;
\While{$|\sigma_1| < \tau$}{
    \If{$|\sigma_1|=\tau-1$}{
        $E'=\{e_{ij}| v_j \notin \sigma_1) \land (v_i=v_j \vee v_i \in \sigma_1)\}$\;
        $e_{ij}=\arg\max_{e_{ij} \in E'} h(e|E(\sigma_1))$\;
        $\sigma_1 = \sigma_1 \oplus v_j$\;
    }
    \Else{
        $e_{ij}=\arg\max_{\{e_{ij}|v_j\notin \sigma_1\}} h(e|E(\sigma_1))$\;
        \If{$v_j=v_i$ or $v_i\in \sigma_1$}{
            $\sigma_1 = \sigma_1 \oplus v_j$\;
        \Else{
            $\sigma_1 = \sigma_1 \oplus v_i \oplus v_j$\;
            }
        }
    }
}
\While{$|\sigma_2| < k-\tau$}{
    \If{$|\sigma_2|=k-\tau-1$}{
        $E'=\{e_{ij}| v_j \notin \sigma_1 \cup \sigma_2 \land (v_i=v_j \vee v_i \in \sigma_2)\}$\;
        $e_{ij}=\arg\max_{e_{ij}\in E'} h(e|E(\sigma_2))$\;
        $\sigma_2 = \sigma_2 \oplus v_j$\;
    }
    \Else{
        $e_{ij}=\arg\max_{\{e_{ij}|v_i\notin \sigma_1, v_j \notin \sigma_1 \cup \sigma_2\}} h(e|E(\sigma_2))$\;
        \If{$v_j=v_i$ or $v_i\in \sigma_2$}{
            $\sigma_2 = \sigma_2 \oplus v_j$\;
        \Else{
            $\sigma_2 = \sigma_2 \oplus v_i \oplus v_j$\;
            }
        }
    }
}
$\sigma=\sigma_1\oplus \sigma_2$\;
\Return $\sigma$
\end{algorithm}

\subsection{Theoretical Results}
Let $\alpha=2 d_{\text{in}}+1$, $\beta=1+d_{\text{in}}+d_{\text{out}}$, $\gamma=e^{\frac{k-3}{k-2}}$, $\eta=e^{\frac{k-2\tau-1}{k-\tau-1}}$. Note $d_{\text{in}}$ and $d_{\text{out}}$ are the maximum in and out degree of the network respectively. For convience, we denote $f(v|\sigma)$ and $f(\sigma'|\sigma)$ as the marginal gain of attending $v$ and $\sigma'$ to sequence $\sigma$ respectively. We denote $\sigma^*(V,k,\tau)$ as the optimal solution of the RoseNets problem with element set $V$, cardinality $k$ and robust parameter $\tau$, and $g_\tau(\sigma)$ be the minimum value of $f(\sigma)$ after $\tau$ elements are removed from $\sigma$.

\begin{theorem}
Consider $\tau=1$, Algorithm 1 achieves an approximation ratio of \[\max\{\frac{1-e^{-(1-1/k)}}{\alpha\beta},\frac{\gamma^{\frac{1}{d_{\text{in}}}}-1}{\beta \gamma^{\frac{1}{d_{\text{in}}}}-1}\}.\]
\end{theorem}

\begin{theorem}
Consider $1\le \tau \le k$, Algorithm 1 achieves an approximation ratio of \[\max\{\frac{1-e^{-(1-1/k)}}{\alpha\beta},\frac{\tau\alpha\beta(\eta^{\frac{1}{d_{\text{in}}}}-1)}{\tau\alpha\eta^{\frac{1}{d_{\text{in}}}}- \beta (1-e^{-(1-1/k)}) }\}.\]
\end{theorem}

In Theorem 1, it is hard to compare the two approximation ratios directly due to their complex mathematical expression. Thus we consider specific network setting to show the different advantages of the two terms.

First, it is easy to verify that both the two terms are monotonically increasing function of $k$. When $k=3$, we have $\frac{1-e^{-(1-1/k)}}{\alpha\beta}-\frac{\gamma^{\frac{1}{d_{\text{in}}}}-1}{\beta \gamma^{\frac{1}{d_{\text{in}}}}-1} = \frac{1-e^{-2/3}}{\alpha \beta}>0$. Thus we know that when $k$ is small, the first term is larger. When $k \to \infty$, the first term has a limit value of $ (1-1/e)/\alpha\beta$, and the second term has a limit value of $\frac{e^{\frac{1}{d_{\text{in}}}}-1}{\beta e^{\frac{1}{d_{\text{in}}}}-1}$.When $d_{\text{in}}=1$, then $\alpha=3$ and  $(1-1/e)/\alpha\beta-\frac{e^{\frac{1}{d_{\text{in}}}}-1}{\beta e^{\frac{1}{d_{\text{in}}}}-1} =(1-1/e)/3\beta-\frac{e-1}{\beta e-1}=\frac{-1-2\beta e +\frac{1}{e}+2\beta}{3\beta(\beta e-1)}<0$. In this case, the second term is larger than the first term in Theorem 1. Thus we can conclude that under specific network structure, the second term would be larger with large $k$.

Similarly, for Theorem 2, when $k=3$ and $\tau=1$, $\frac{1-e^{-(1-1/k)}}{\alpha\beta}-\frac{\tau\alpha\beta(\eta^{\frac{1}{d_{\text{in}}}}-1)}{\tau\alpha\eta^{\frac{1}{d_{\text{in}}}}- \beta (1-e^{-(1-1/k)})} = \frac{1-e^{-2/3}}{\alpha \beta}>0$. Thus we know when $k$ and $\tau$ is small, the first term is larger. When $k \to \infty$ while $\tau$ remains a constant, the first term has a limit value of $ (1-1/e)/\alpha\beta$, the second term has a limit value of $\frac{\tau\alpha\beta(e^{\frac{1}{d_{\text{in}}}}-1)}{\tau\alpha e^{\frac{1}{d_{\text{in}}}}- \beta (1-e^{-1}) }$. When $d_{\text{in}}=1$ and $d_{\text{out}}<\frac{3\tau e^2}{e-1}-2$, then $\alpha=3$, $\beta < \frac{3\tau e^2}{e-1}$ and $(1-1/e)/\alpha\beta-\frac{\tau\alpha\beta(e^{\frac{1}{d_{\text{in}}}}-1)}{\tau\alpha e^{\frac{1}{d_{\text{in}}}}- \beta (1-e^{-1}) } = \frac{(1-\frac{1}{e})(3\tau e -\beta(1-\frac{1}{e}))-9\beta^2 \tau(e-1)}{3\beta(3\tau e -\beta(1-\frac{1}{e}))}<0$. In this case, the second term is larger than the first term in Theorem 2. Thus we can conclude that under specific network structure, the second term would be larger with large $k$.

According to the above analysis, we know that the value of $k$ and $\tau$, together with the network topology, significantly affect the approximation ratio. It would be an interesting future direction to explore how the approximation ratio change when the parameters and network topology change.

To prove the above two theorems, we need the following three auxiliary lemmas. Due to space limitations, we here assume Lemma 1, 2 and 3 hold and show the proof of Theorem 1 below. We provide the proofs of Lemma 1, 2, 3 and Theorem 2 in the supplementary material.

\begin{lemma}
There exists an element $v$ for sequence $\sigma_1$ and $\sigma_2$ satisfies that $f(v|\sigma_1) \ge \frac{1}{d_{\text{in}}|\sigma_2|} f(\sigma_2|\sigma_1)$.
\end{lemma}

\begin{lemma}
Consider $c \in (0,1]$ and $1 \le k' \le k$. Suppose that the sequence selected is $\sigma$ with $|\sigma| = k$ and that there exists a sequence $\sigma'$ with $|\sigma'| = k-k'$ such that $\sigma' \subseteq \sigma$ and $f(\sigma') \ge c f(\sigma)$. Then we have $f(\sigma) \ge \frac{e^{\frac{k'}{d_{\text{in}}k}}-1} {e^{\frac{k'}{d_{\text{in}}k}}-c} f(\sigma^*(V,k,0))$.
\end{lemma}

\begin{lemma}
Consider $1 \le \tau \le k$. The following holds for any $Z \subseteq V$ with $|Z| \le \tau$: $g_\tau (\sigma^*(V,k,\tau)) \le f(\sigma^*(V-Z,k-\tau,0))$.
\end{lemma}

\subsection{Proof of Theorem 1}
Given $\tau=1$, the selected sequence $\sigma_1$ has one element. And we have $\sigma_2=k-1$. Let $\sigma_1=\{v_1\}$. Then the final sequence is $\sigma= \{v_1\} \oplus \sigma_2$. Suppose the the removed vertex from the sequence is $z$.

First, we show a lower bound on $f(\sigma_2)$:
\begin{equation}
\begin{aligned}
f(\sigma_2) & \ge \frac{1-e^{-(1-1/k)}}{\alpha} f(\sigma^*(V\backslash v_1,k-1,0)) \\
& \ge \frac{1-e^{-(1-1/k)}}{\alpha} g_\tau(\sigma^*(V,k,\tau))
\end{aligned}
\end{equation}
The first inequality is due to the approximation ratio of Sequence Greedy algorithm for net-submodular maximization \cite{mitrovic2018submodularity}. The second inequality is due to Lemma 3.

Now, we can see the removed element $z$ can be either $v_1$ or an element in $\sigma_2$. In the following, we will consider these two cases:

\textbf{Case 1.} Let $z=v_1$. Then we have
\begin{equation}
f(\sigma -z ) = f(\sigma_2) \ge \frac{1-e^{-(1-1/k)}}{\alpha} g_\tau(\sigma^*(V,k,\tau))
\end{equation}

\textbf{Case 2.} Let $z\in \sigma_2$. We then further consider two cases:

\textbf{Case 2.1.} Let $f(\sigma_2) \le f(\sigma_2-z)$.

In this case, the removal does not reduce the overall value of the remaining sequence $\sigma_2-\{z\}$. Then we have
\begin{equation}
\begin{aligned}
f(\sigma-v) & =f(v_1 \oplus (\sigma_2-z)) \ge f(\sigma_2-z) \\
& \ge f(\sigma_2) \ge \frac{1-e^{-(1-1/k)}}{\alpha} g_\tau(\sigma^*(V,k,\tau))
\end{aligned}
\end{equation}

\textbf{Case 2.2.} Let $f(\sigma_2) > f(\sigma_2-z)$.

We define $q= \frac{f(\sigma_2)-f(\sigma_2-z)}{(d_{\text{in}}+d_{\text{out}})f(\sigma_2)}$, to represent the ratio of the loss of removing element $z$ from sequence $\sigma_2$ to the value of the sequence $\sigma_2$. Obviously, we have $q\in (0,\frac{1}{d_{\text{in}}+d_{\text{out}}}]$ since $f(\sigma_2) > f(\sigma_2-z)$.

First, we have
\begin{equation}
\begin{aligned}
(d_{\text{in}} + & d_{\text{out}}) q f(\sigma_2)= f(\sigma_2) - f(\sigma_2-z) \\
& = f(\sigma_2^1 \oplus z \oplus \sigma_2^2) - f(\sigma_2^1 \oplus \sigma_2^2) \\
& = f(\sigma_2^1) + f(z|\sigma_2^1)+f(\sigma_2^2| (\sigma_2^1\oplus z)) \\
 & \quad\quad\quad\quad\quad\quad - f(\sigma_2^1) -f(\sigma_2^2|\sigma_2^1) \\
& = f(z|\sigma_2^1)+f(\sigma_2^2| (\sigma_2^1\oplus z)) - f(\sigma_2^2|\sigma_2^1) \\
& \le d_{\text{in}} h(e_z^{\text{in}}) + d_{\text{out}} h(e_z^{\text{out}}) \\
& \le (d_{\text{in}} + d_{\text{out}}) \max \{h(e_z^{\text{in}}),h(e_z^{\text{out}})\}
\end{aligned}
\end{equation}
where $h(e_z^{\text{in}})$/$h(e_z^{\text{out}})$ are the edge that has maximum utility over all the incoming/outgoing edges of $z$. The first inequality is due to the fact that the marginal gain of a vertex $z$ to the prefix and subsequent sequence is at most $d_{\text{in}} h(e_z^{\text{in}})$ and $d_{\text{out}} h(e_z^{\text{out}})$. Then the second inequality follows intuitively.

Given Equation (4), we need to prove four inequalities for finally proving the theorem.

First, suppose the first vertex of $\sigma_2$ is $v_2$. By the monotonicity of function $f(\cdot)$ and Equation (4), we have
\begin{equation}
\begin{aligned}
f(\sigma-\{z\}) & \ge f(v_1 \oplus v_2) \\
& \ge \max \{h(e_z^{\text{in}}),h(e_z^{\text{out}})\} \ge q f(\sigma_2), \text{ and } \\
f(\sigma-\{z\}) & = f(v_1 \oplus (\sigma_2-z)) \\
& \ge f(\sigma_2 -z) \ge (1-(d_{\text{in}} + d_{\text{out}})q) f(\sigma_2)
\end{aligned}
\end{equation}
Given Equation (5), we have Inequality 1 as below.

\textbf{Inequality 1:}
\[
f(\sigma-z)\ge \max\{ q \cdot f(\sigma_2),(1-(d_{\text{in}} + d_{\text{out}})q) \cdot f(\sigma_2)\}.
\]

We know $\max\{x,1-bx\} \ge \frac{1}{1+b}$ for $x\in (0,\frac{1}{b}]$ and $b>0$.\footnote{As $x$ is monotone increasing and $1-bx$ is monotone decreasing, the function achieves the maximum value when $x=1-bx$} Thus we have Inequality 2 as below.

\textbf{Inequality 2: }$\max\{q,1-(d_{\text{in}} + d_{\text{out}})q\} \ge 1/\beta.$

Note that the first two elements $v_2,v_3$ in $\sigma_2$ satisfy that
\[
f(v_2\oplus v_3) \ge \max \{h(e_z^{\text{in}}),h(e_z^{\text{out}})\} \ge q f(\sigma_2)
\]

Thus by replacing the parameters in Lemma 2 and Lemma 3, we have the following result, which implies Inequality 3.
\[
\begin{aligned}
& f(\sigma_2) \ge \frac{\gamma^{\frac{1}{d_{\text{in}}}}-1}{\gamma^{\frac{1}{d_{\text{in}}}}-q} f(\sigma^*(V\backslash v_1,k-\tau,0)) \\
& \Longrightarrow \text{ \textbf{Inequality 3: }}f(\sigma_2) \ge \frac{\gamma^{\frac{1}{d_{\text{in}}}}-1}{\gamma^{\frac{1}{d_{\text{in}}}}-q} g_{\tau} (\sigma^*(V,k,\tau))
\end{aligned}
\]

Now define $\ell_1(q) = q \frac{\gamma^{\frac{1}{d_{\text{in}}}}-1}{\gamma^{\frac{1}{d_{\text{in}}}}-q}$ and $\ell_2(q) = (1-(d_{\text{in}} + d_{\text{out}})q) \frac{\gamma^{\frac{1}{d_{\text{in}}}}-1}{\gamma^{\frac{1}{d_{\text{in}}}}-q} \} \ge \frac{\gamma^{\frac{1}{d_{\text{in}}}}-1}{\beta \gamma^{\frac{1}{d_{\text{in}}}}-1}
$. It is easy to verify that for $k\ge 3$ and $q\in (0,\frac{1}{d_{\text{in}}+d_{\text{out}}}]$, $\ell_1(q)$/$\ell_2(q)$ is monotonically increasing/decreasing. Note when $q=\frac{1}{\beta}$, $\ell_1(q)=\ell_2(q)=\frac{\gamma^{\frac{1}{d_{\text{in}}}}-1}{\beta \gamma^{\frac{1}{d_{\text{in}}}}-1}$. We consider two cases for $q$: (1) when $q \in (0,\frac{1}{\beta}]$, we have $\max\{\ell_1(q),\ell_2(q)\}\ge \ell_2(\frac{1}{\beta})$  as $\ell_2(q)$ is monotonically decreasing; (2) when $q \in (\frac{1}{\beta},\frac{1}{d_{\text{in}}+\text{out}}]$, we have $\max\{\ell_1(q),\ell_2(q)\}\ge \ell_1(\frac{1}{\beta})$ as $\ell_1(q)$ is monotonically increasing. Thus we have the Inequality 4 as below.

\textbf{Inequality 4:}
\[
\max\{ q \frac{\gamma^{\frac{1}{d_{\text{in}}}}-1}{\gamma^{\frac{1}{d_{\text{in}}}}-q} , (1-(d_{\text{in}} + d_{\text{out}})q) \frac{\gamma^{\frac{1}{d_{\text{in}}}}-1}{\gamma^{\frac{1}{d_{\text{in}}}}-q} \} \ge \frac{\gamma^{\frac{1}{d_{\text{in}}}}-1}{\beta \gamma^{\frac{1}{d_{\text{in}}}}-1}
\]

Combining Inequality 1, Inequality 2 and Equation (1), we can have the first lower bound in Theorem 1:
\[
\begin{aligned}
& f(\sigma -z )\ge \max\{ q \cdot f(\sigma_2),(1-(d_{\text{in}} + d_{\text{out}})q) f(\sigma_2)\} \\
\ge & \max\{q,1-(d_{\text{in}} + d_{\text{out}})q\} \frac{1-e^{-(1-1/k)}}{\alpha} g_\tau(\sigma^*(V,k,\tau)) \\
\ge & \frac{1-e^{-(1-1/k)}}{\alpha\beta} g_\tau(\sigma^*(V,k,\tau))
\end{aligned}
\]

Combining Inequality 1, Inequality 3 and Inequality 4, we can have the second lower bound in Theorem 1:
\[
\begin{aligned}
& f(\sigma -z ) \ge \max\{ q \cdot f(\sigma_2),(1-(d_{\text{in}} + d_{\text{out}})q) \cdot f(\sigma_2)\} \\
& \ge \max\{ q \cdot \frac{\gamma^{\frac{1}{d_{\text{in}}}}-1}{\gamma^{\frac{1}{d_{\text{in}}}}-q} , \\
& \quad\quad\quad (1-(d_{\text{in}} + d_{\text{out}})q) \frac{\gamma^{\frac{1}{d_{\text{in}}}}-1}{\gamma^{\frac{1}{d_{\text{in}}}}-q} \} g_\tau(\sigma^*(V,k,\tau)) \\
& \ge \frac{\gamma^{\frac{1}{d_{\text{in}}}}-1}{\beta \gamma^{\frac{1}{d_{\text{in}}}}-1} g_\tau(\sigma^*(V,k,\tau))
\end{aligned}
\]
Now we are done.

\section{Experiments}
We compare the performance of our algorithms RoseNets to the non-robust version Sequence Greedy (\textbf{Sequence} for short) \cite{mitrovic2018submodularity}, the existing submodular sequence baseline (\textbf{OMegA}) \cite{tschiatschek2017selecting}, and a naive baseline (\textbf{Frequency}) which outputs the most popular items the user has not yet reviewed.

To evaluate the performance of the algorithms, we use three evaluation metrics in this paper. The first one is \textbf{Accuracy Score}, which simply counts the number of accurately recommended items. While this is a sensible measure, it does not explicitly consider the order of the sequence. Therefore, we also consider the \textbf{Sequence Score}, which is a measure based on the Kendall-Tau distance \cite{kendall1938new}. This metric counts the number of ordered pairs that appear in both the predicted sequence and the true sequence. These two metrics are also used in \cite{mitrovic2019adaptive}. The third metric is the \textbf{Utility Function Value} of the selected sequence.

We use a probabilistic coverage utility function as our Net-submodular function $f$. Mathematically,
\[
f(\sigma)=h(E_1)=\sum_{j\in V} [1- \prod_{(i,j)\in E_1} (1-w_{ij})],
\]
where $E_1\in E$ is the edges induced by sequence $\sigma$. And to simulate the worst case removal, we remove the first $\tau$ elements to evaluate the robustness of the algorithms.

\subsection{Amazon Product Recommendation}
Using the Amazon Video Games review dataset \cite{ni2019justifying}, we conduct experiments for the task of recommending products to users. In particular, given a specific user with the first 4 products she has purchased, we want to predict the next $k$ products she will buy. We first build a graph $G = (V,E)$, where $V$ is the set of all products and $E$ is the set of edges between these products. The weight of each edge, $w_{ij}$, is defined to be the conditional probability of purchasing product $j$ given that the user has previously purchased product $i$. We compute $w_{ij}$ by taking the fraction of users that purchased $j$ after having purchased $i$ among all the users that purchased $i$. There are also self-loops with weight $w_{ii}$ that represent the fraction of users that purchased product $i$ among all the users. We focused on the products that have been purchased at least 50 times each, leaving us with a total of 9383 unique products. Also we select the users that have purchased at least 29 products, leaving use 909 users. We conduct recommendation task on these 909 users and take the average value of each evaluation metric.

Figure 3 shows the performance of the comparison algorithms using the accuracy score, sequence score and utility function value respectively. In Figure 3(a), 3(b), 3(c) and 3(d), we find that after the removal of $\tau$ elements, the RoseNets outperforms all the comparisons. Such results demonstrate that the RoseNets algorithm is effective and robust in real applications since accuracy score and sequence score are common evaluation metrics in practical. In Figure 3(e) and 3(f), the only difference is that the OMegA algorithm is outperforming when $\tau$ is small or $k$ is large. The OMegA algorithm aims to find a global optimal solution. It topologically resorts all the candidates after each element selection. It can return a solution with better utility function value when $k$ is large and $\tau$ is small, but runs much slower than RoseNets or Sequence. Also, it shows poor performance in accuracy and sequence score. Thus the RoseNets algorithm is more effective and robust in real applications.

In addition, we also show the case of RoseNets and Sequence with $\tau=0$. We can see that in all the experimental results for the three metrics, the RoseNets outperforms Sequence, which is consistent to our expectation. On utility function value, the Sequence($\tau=0$) is better than the RoseNets($\tau=0$). However, on accuracy score and sequence score, the RoseNets($\tau=0$) is very close to the Sequence($\tau=0$), sometimes shows better performance. The former result is due to the effectiveness of greedy framework. The RoseNets algorithm indeed invokes Sequence for two trials independently, which intuitively cannot achieve comparable performance with one trial Sequence execution, due to the Net-submodularity. But the latter result shows that directly implementing greedy selection is not always outperforming. This is due to the intrinsic property of the greedy algorithm. Though $1-1/e$ is almost the best approximation ratio, some heuristic algorithms may achieve better performance in specific cases. However, if we invoke more trials of Sequence algorithm, better robust experimental results and approximation ratio might be achieved but the utility value would become lower. This is because more trials of independent greedy selection would give high probability of triggering the diminishing return phenomenon. In real applications, this is a trade-off for designing robust algorithms that requires a balance between high efficiency on robustness and maximization of utility value.

\begin{figure}[t]
  \centering
  \subfigure[\small{$\tau=10$ and $k=[11,20]$}]{
    \includegraphics[height=1.24in]{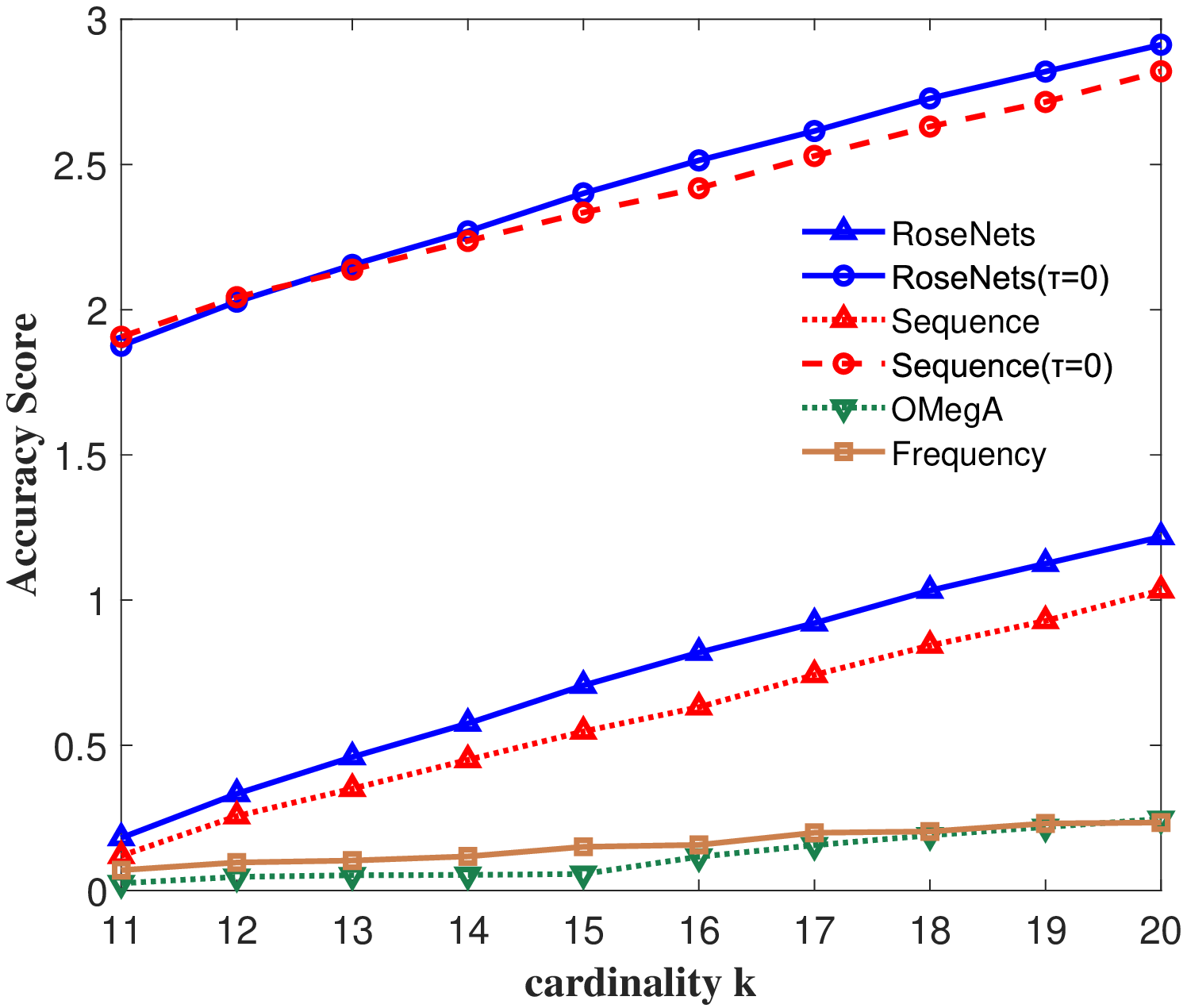}}
  \hspace{-0.5em}
  \subfigure[\small{$k=15$ and $\tau=[0,10]$}]{
    \includegraphics[height=1.24in]{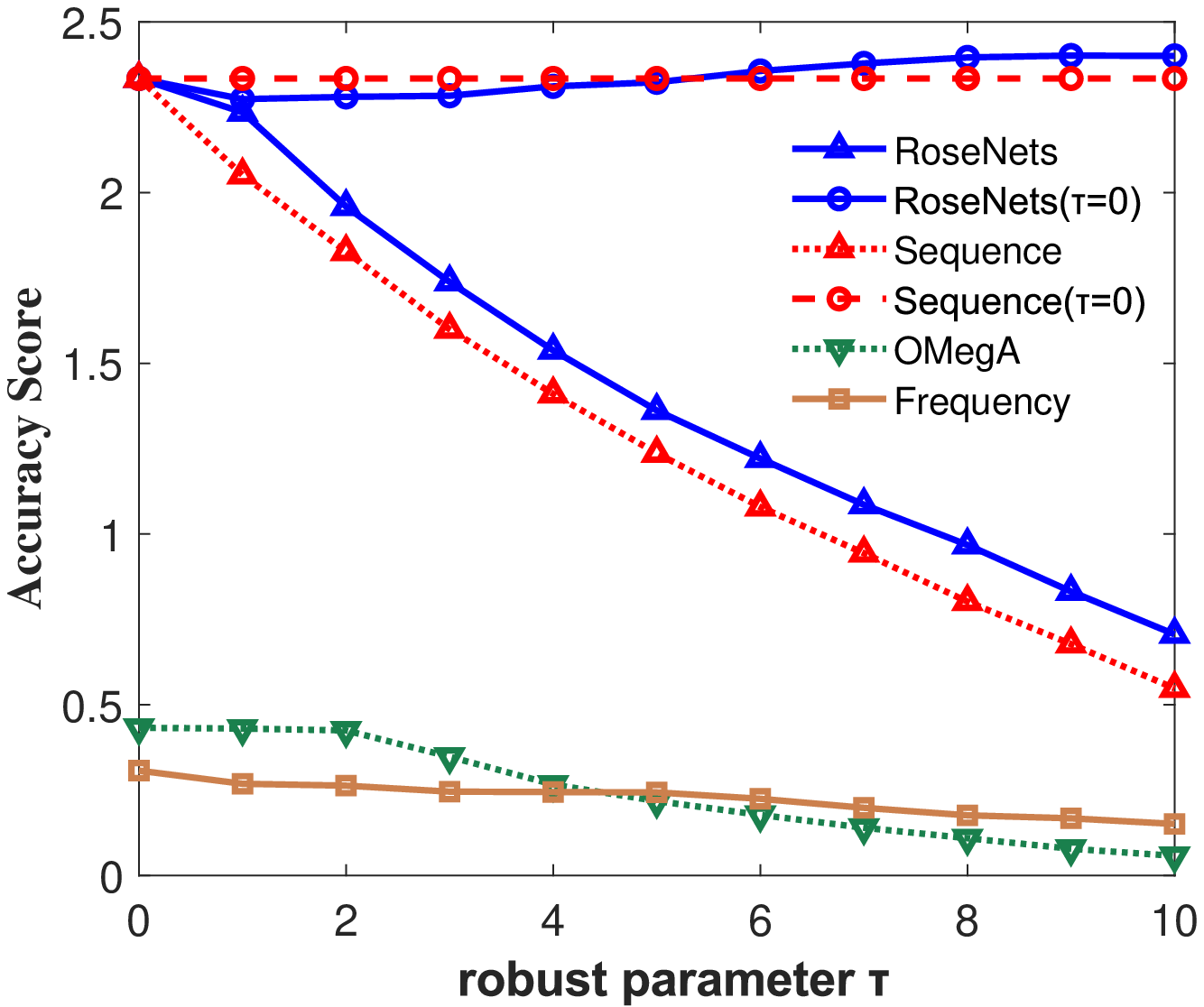}}
  \hspace{-0.5em}
  \subfigure[\small{$\tau=10$ and $k=[11,20]$}]{
    \includegraphics[height=1.24in]{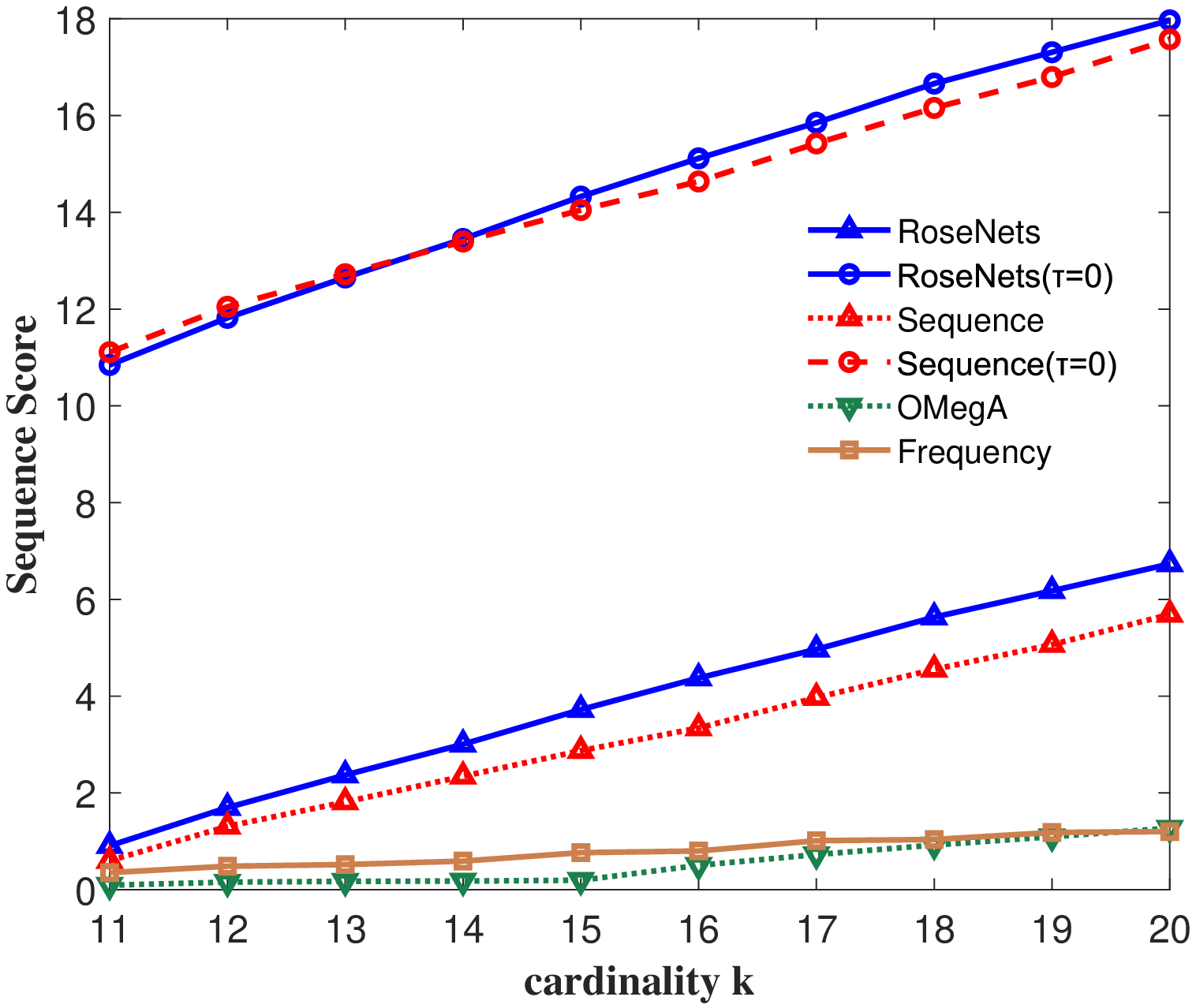}}
  \hspace{-0.5em}
  \subfigure[\small{$k=15$ and $\tau=[0,10]$}]{
    \includegraphics[height=1.24in]{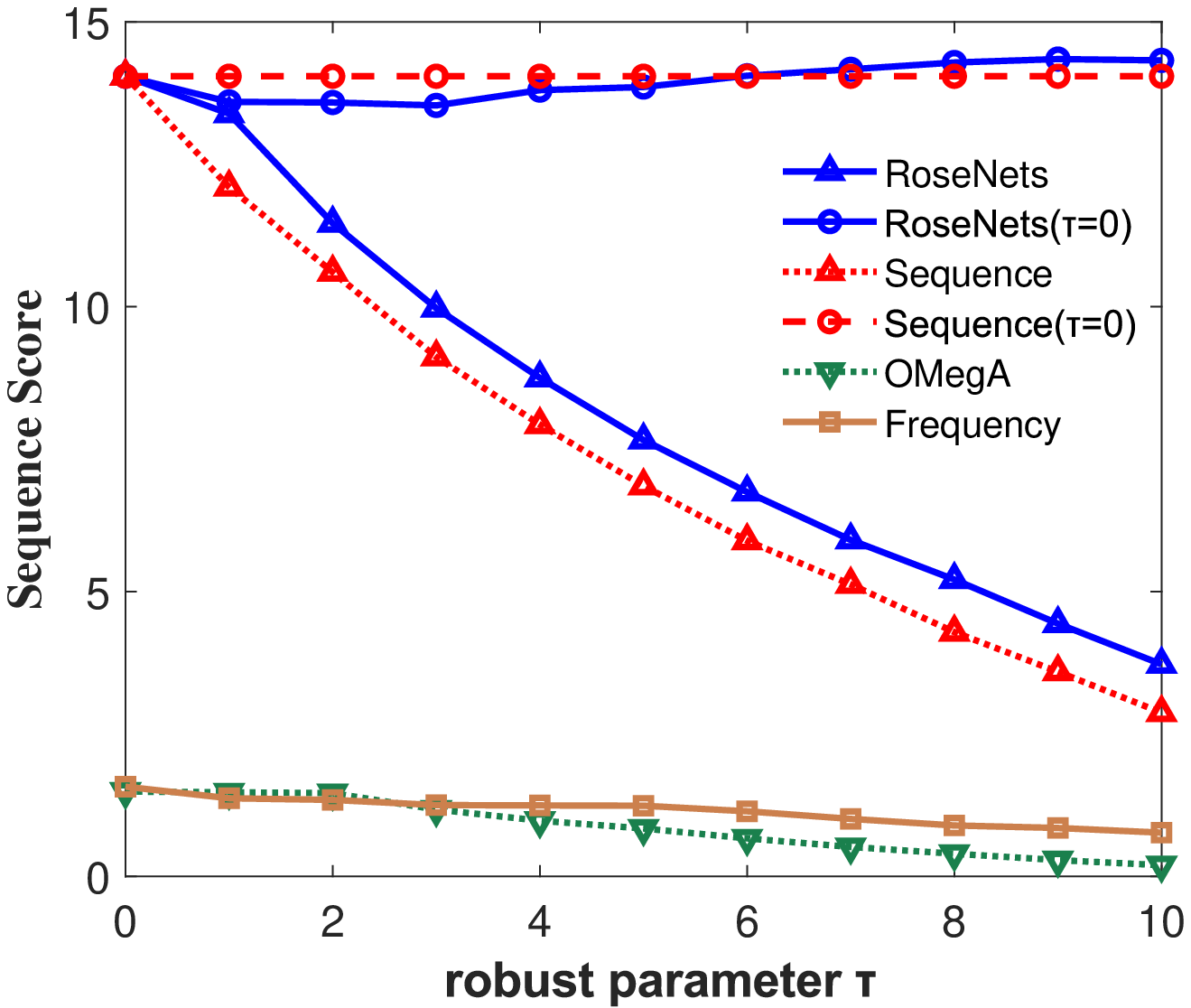}}
  \hspace{-0.5em}
  \subfigure[\small{$\tau=10$ and $k=[11,20]$}]{
    \includegraphics[height=1.24in]{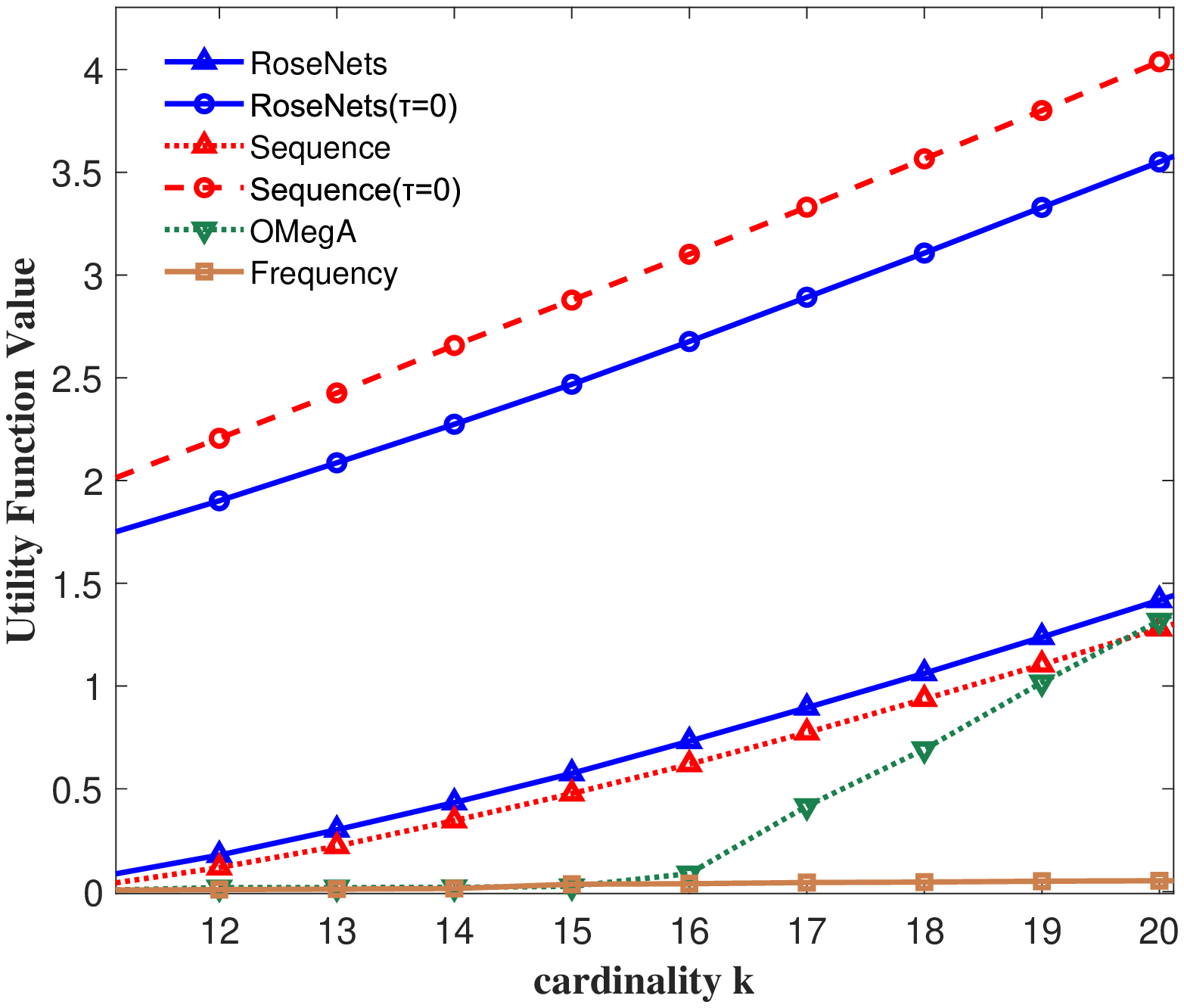}}
  \hspace{-0.5em}
  \subfigure[\small{$k=15$ and $\tau=[0,10]$}]{
    \includegraphics[height=1.24in]{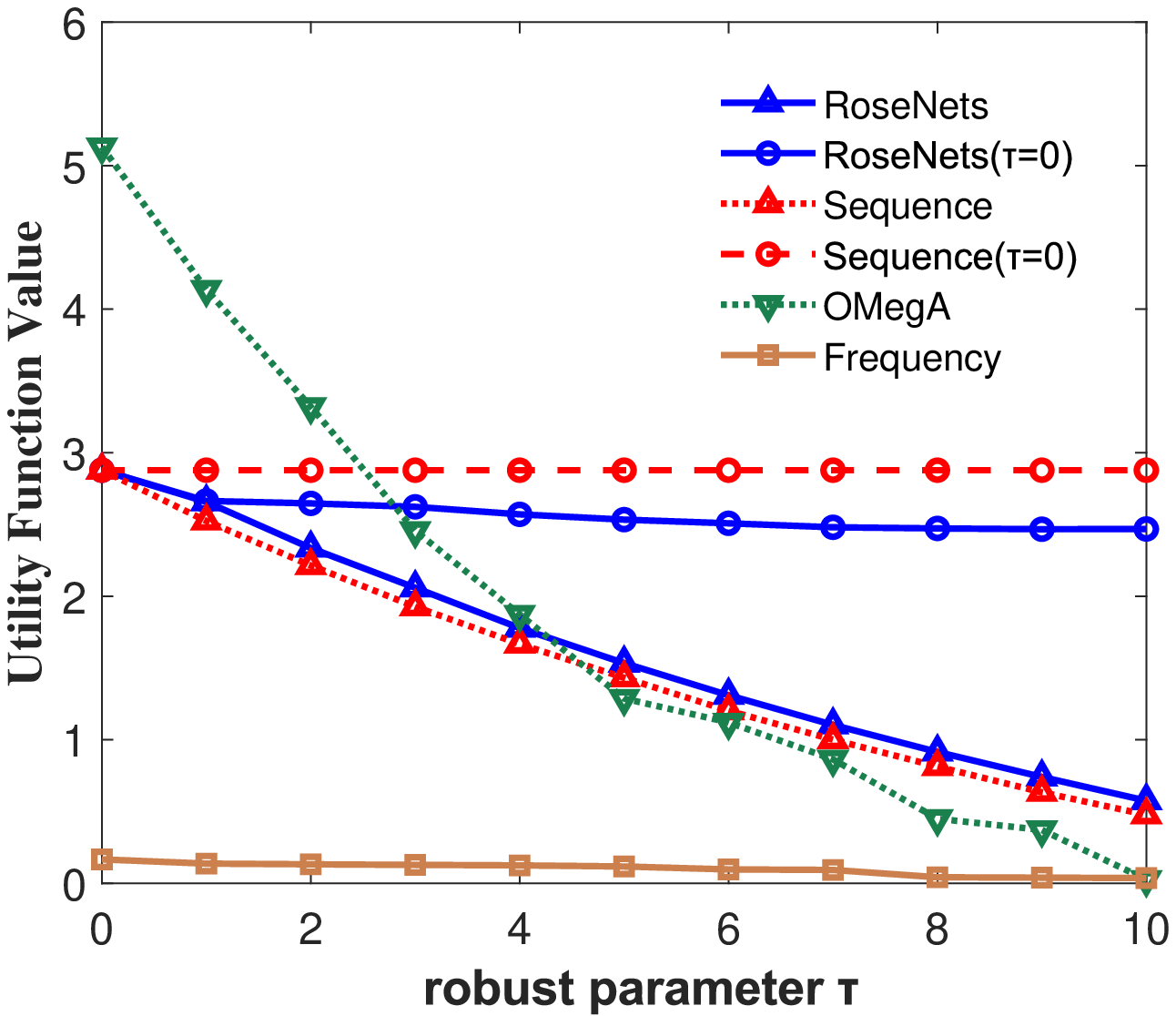}}
  \caption{Recommendation Application}
\vspace{-1.5em}
\end{figure}

\begin{figure}[t]
  \centering
  \subfigure[\small{$\tau=5$ and $k=[6,15]$}]{
    \includegraphics[height=1.24in]{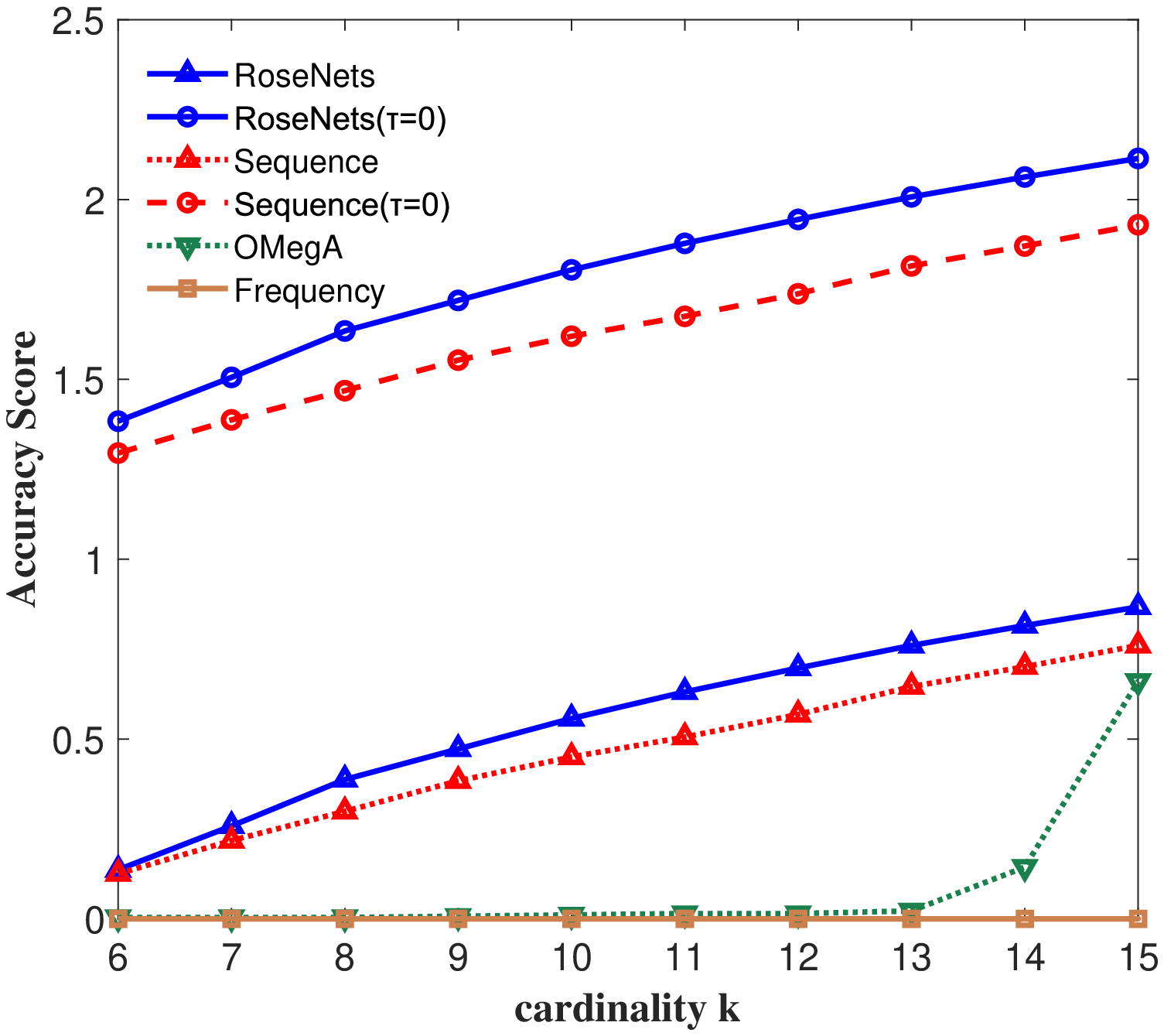}}
  \hspace{-0.5em}
  \subfigure[\small{$k=10$ and $\tau=[0,10]$}]{
    \includegraphics[height=1.24in]{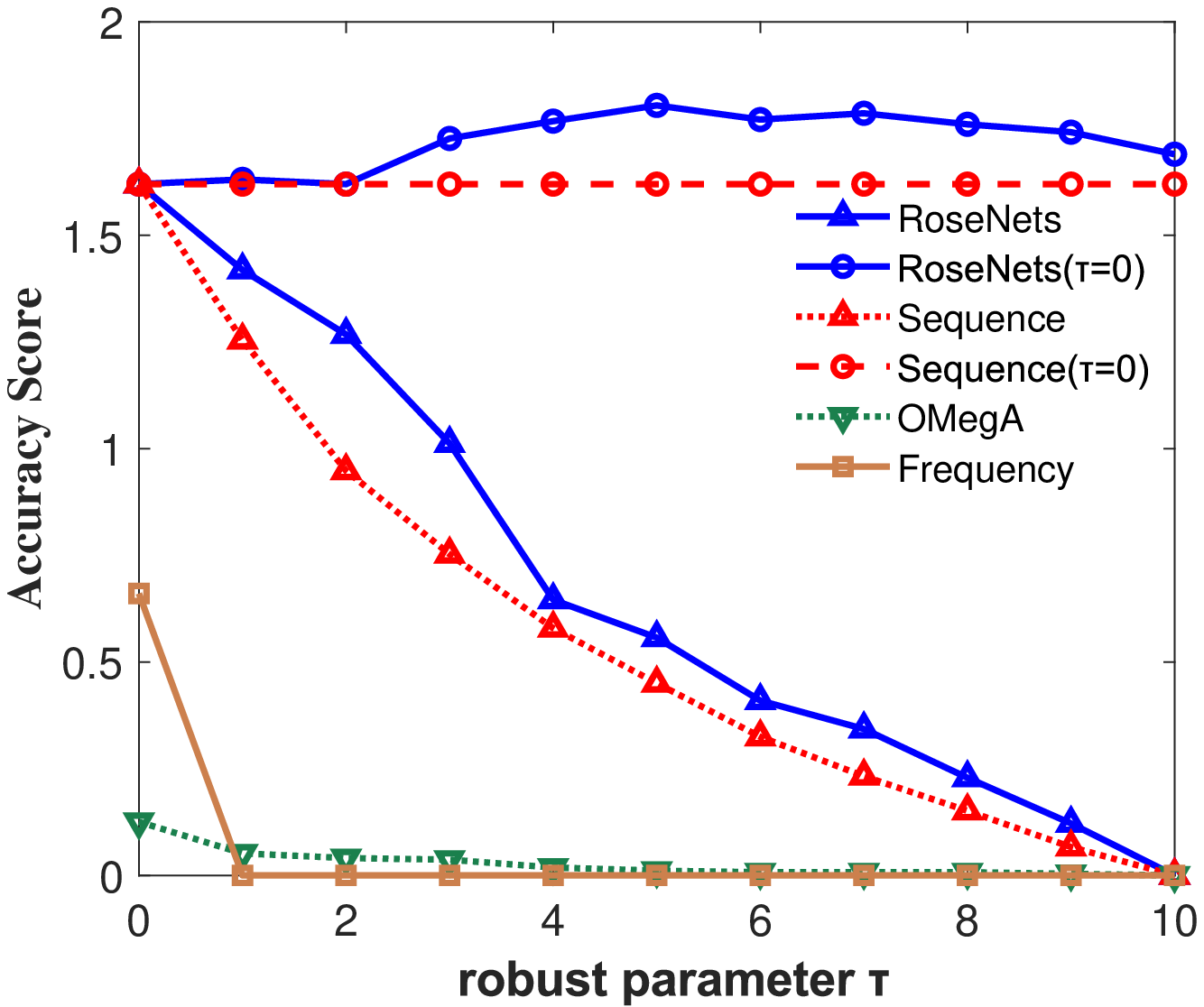}}
  \hspace{-0.5em}
  \subfigure[\small{$\tau=5$ and $k=[6,15]$}]{
    \includegraphics[height=1.24in]{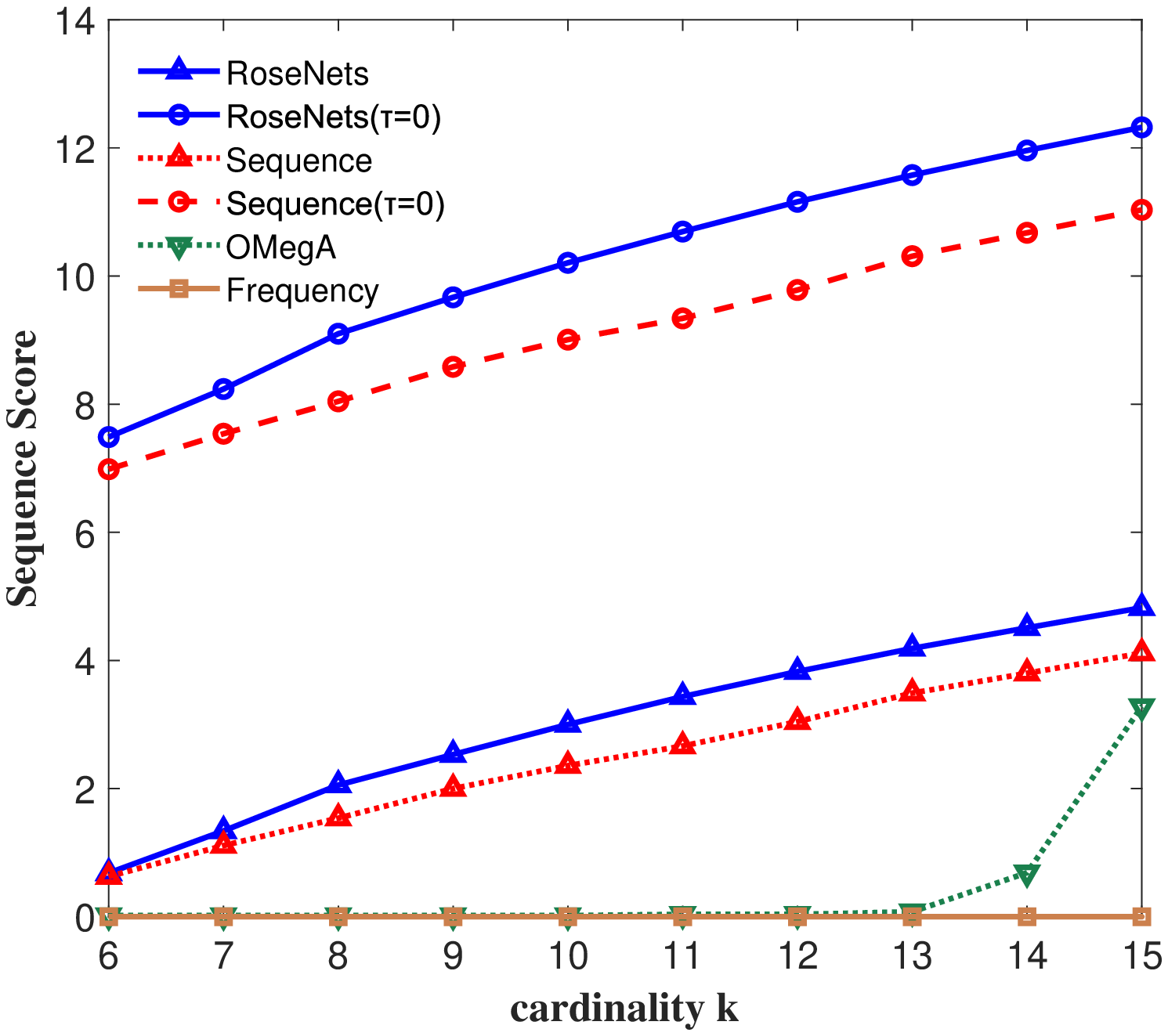}}
  \hspace{-0.5em}
  \subfigure[\small{$k=10$ and $\tau=[0,10]$}]{
    \includegraphics[height=1.24in]{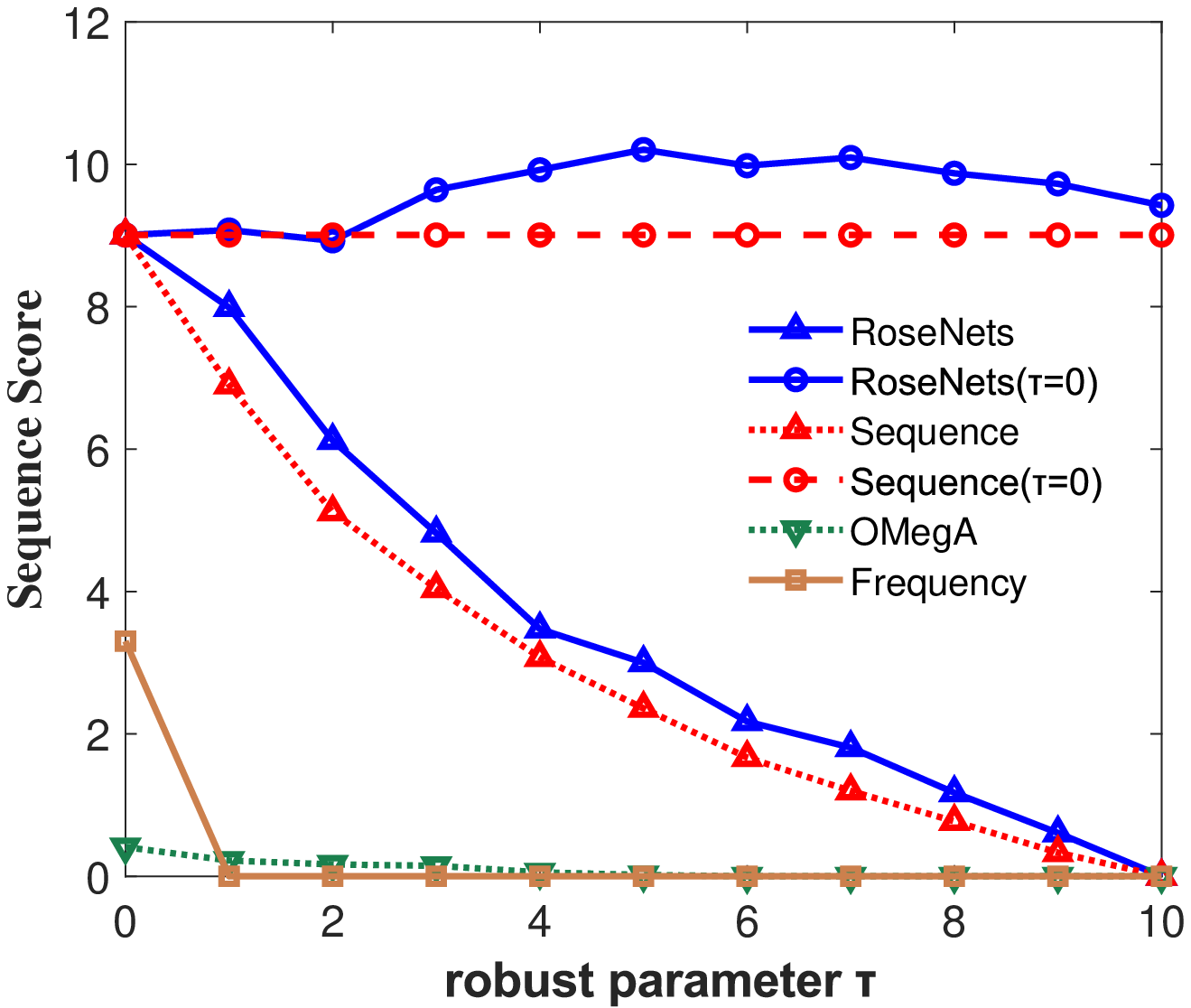}}
  \hspace{-0.5em}
  \subfigure[\small{$\tau=5$ and $k=[6,15]$}]{
    \includegraphics[height=1.24in]{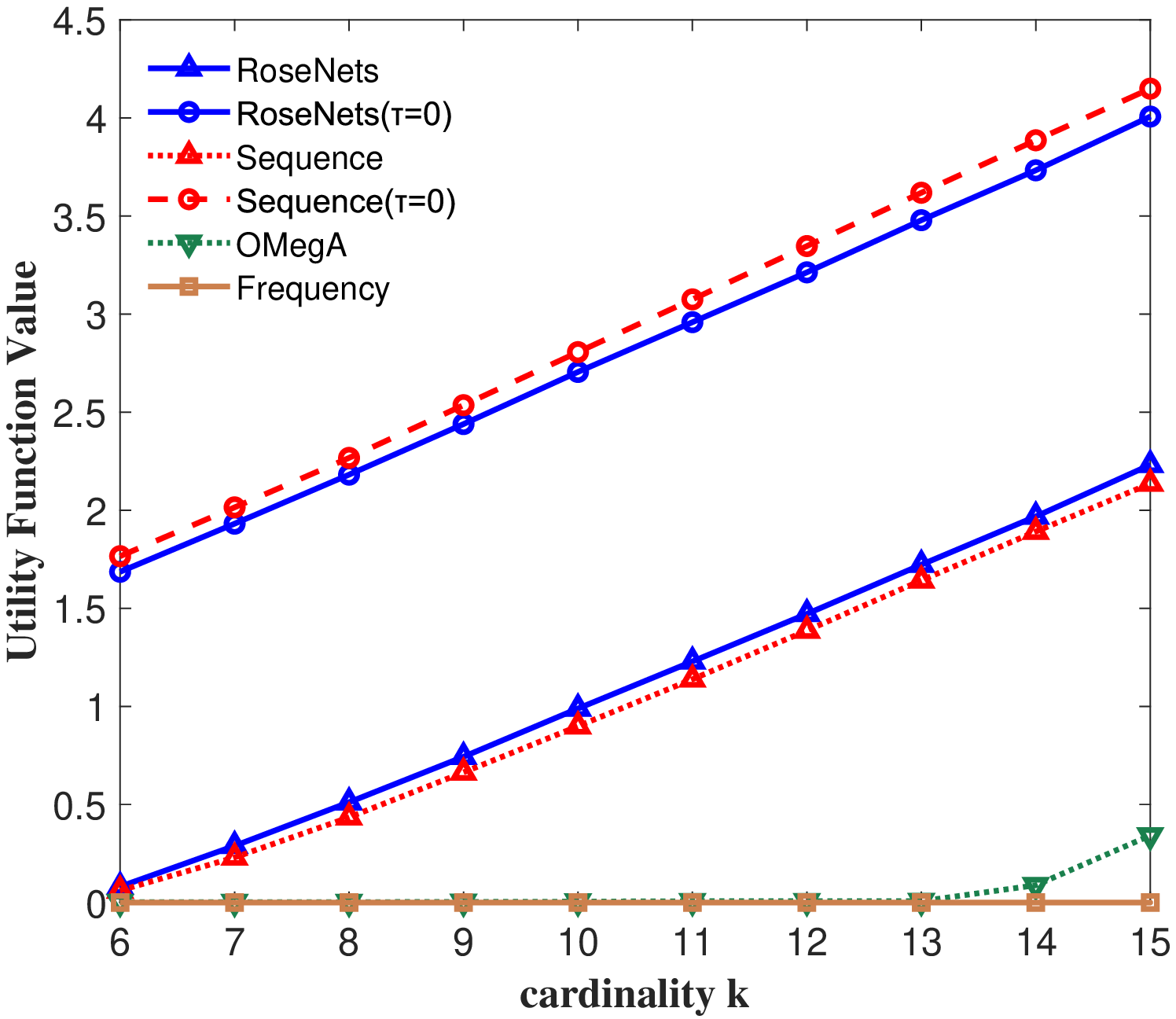}}
  \hspace{-0.5em}
  \subfigure[\small{$k=10$ and $\tau=[0,10]$}]{
    \includegraphics[height=1.24in]{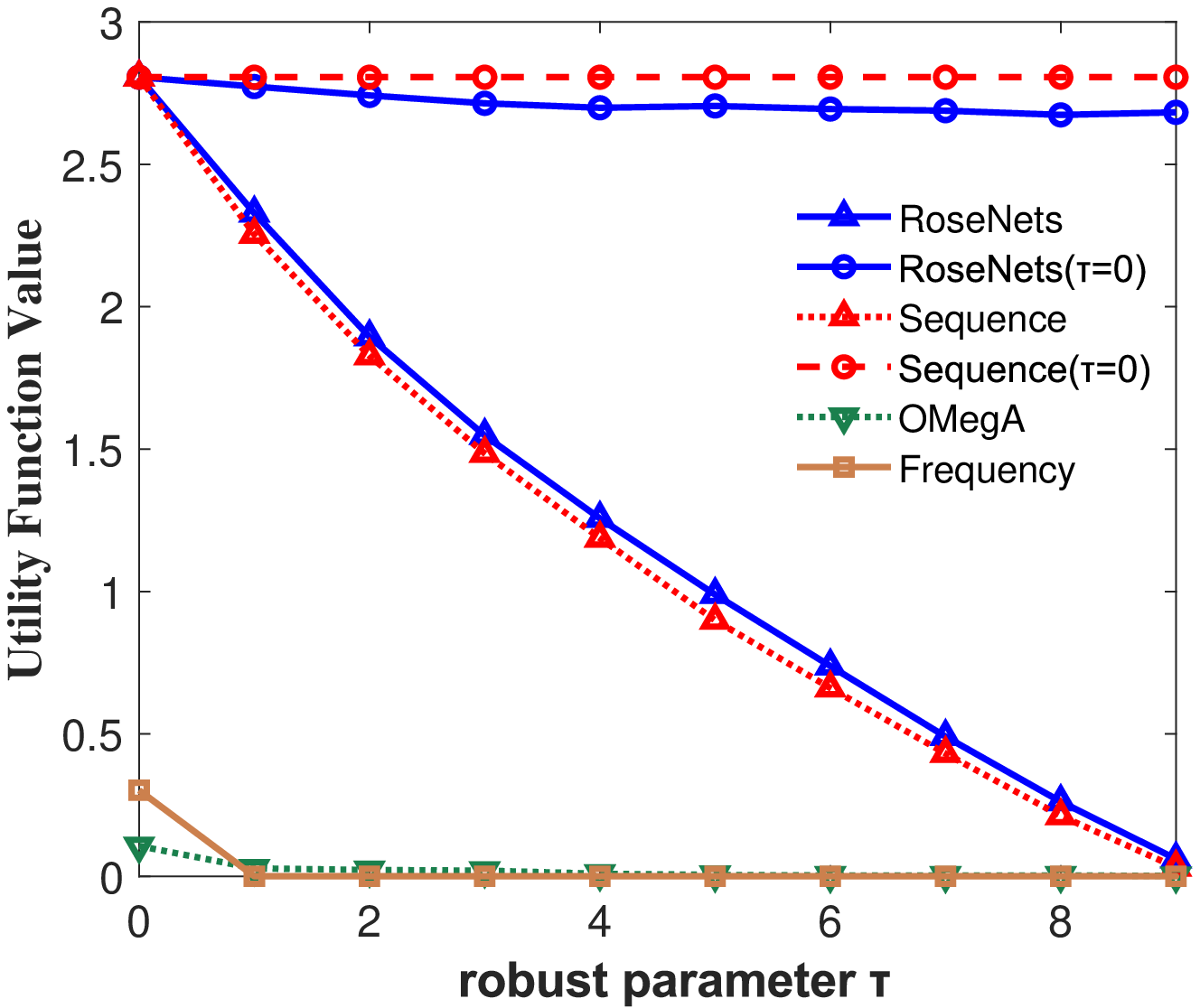}}
  \caption{Link Prediction Application}
\vspace{-1.5em}
\end{figure}

\subsection{Wikipedia Link Prediction}
Using the Wikispeedia dataset \cite{west2009wikispeedia}, we consider users who are surfing through Wikipedia towards some target article. Given a sequence of articles the user has previously visited, we want to guide her to the page she is trying to reach. Since different pages have different valid links, the order of pages we visit is critical to this task. Formally, given the first 4 pages each user visited, we want to predict which page she is trying to reach by making a series of suggestions for which link to follow. In this case, we build the graph $G = (V,E)$, where $V$ is the set of all pages and $E$ is the set of existing links between pages. Similarly to the recommendation case, the weight $w_{ij}$ of an edge $(i,j)\in E$ is the probability of moving to page $j$ given that the user is currently at page $i$, i.e., the fraction of moves from $i$ to $j$ among all the visit of $i$. In this case, we build no self-loops as we assume we can only move using links. Thus we cannot jump to random pages. We condense the dataset to include only articles and edges that appeared in a path, leaving us 4170 unique pages and 55147 edges. We run the algorithm on paths with length at least 29, which leaves us 271 paths. We conduct link prediction task on these 271 paths and take the average value of each evaluation metric.

Figure 4 shows the performance of the comparison algorithms using the accuracy score, sequence score and utility function value respectively. In Figure 4, we find that after the removal of $\tau$ elements, the RoseNets outperforms all the comparisons in all the cases. These results demonstrate that the RoseNets algorithm is effective and robust in real applications. The OMegA algorithm does not show comparable performance to RoseNets algorithm any more. This is because that (1) the path need to be predicted is very long, and (2) the intersection part of different paths is not as large as the case in Amazon recommendation experiments. Thus the global algorithm OMegA cannot exploit its advantages. We still can find in Figure 4(a), 4(c) and 4(e) that when $k$ becomes larger, the performance of OMegA algorithm increases faster. This in turn demonstrates that the RoseNets algorithm is more general, effective and robust, which does not assume specific real application scenario.

Another difference in the link prediction application is that, the RoseNets($\tau=0$) outperforms Sequence($\tau=0$) almost in all the cases on accuracy and sequence score. This again verifies that directly implementing greedy selection may sometimes far away from the optimal solution. As discussed in the end of the recommendation case, an interesting future direction is to explore the trade-off between high efficiency on robustness and the maximization of utility value by invoking more independent greedy selection trials.

\section{Conclusion}
In this paper, we are the first to study the RoseNets problem, which combines the robust optimization and sequence networked submodular maximization. We design a robust algorithm with an approximation ratio that is bounded by the number of the removed elements and the network topology. Experiments on real applications of recommendation and link prediction demonstrate the effectiveness of the proposed algorithm. For future works, one direction is to develop robust algorithms that can achieve higher approximation ratio. An intuitive improvement is to invoke multiple trials of independent greedy selection. Another direction is to consider the robustness against the removal of edges. This is non-trivial since different removal operation would change the network topology and affect the approximation ratio.

\section{Acknowledgments}
This work is supported by the National Natural Science Foundation of China (Grant No: 62102357, U1866602, 62106221, 62102382), and the Starry Night Science Fund of Zhejiang University Shanghai Institute for Advanced Study (Grant No: SN-ZJU-SIAS-001). It is also partially supported by the Zhejiang Provincial Key Research and Development Program of China (2021C01164).

{\small{\bibliography{aaai23}}}

\onecolumn
\section{Appendices: Omitted Proofs}
\subsection{Proof of Lemma 1}
\textbf{Lemma 1.} \textit{
There exists a vertex $v$ for sequence $\sigma_1$ and $\sigma_2$ satisfies that $f(v|\sigma_1) \ge \frac{1}{d_{\text{in}}|\sigma_2|} f(\sigma_2|\sigma_1)$.}
\begin{proof}
Define $E_1=E(\sigma_1)$ and $E_2=E(\sigma_2 \oplus \sigma_1)-E(\sigma_1)$. We denote $E^{[i,j]}$ as the $i$th to $j$th elements in $E$.

\[
\begin{aligned}
f(\sigma_2|\sigma_1) & = f(\sigma_2 \oplus \sigma_1) - f(\sigma_1) = h(E(\sigma_2 \oplus \sigma_1)) - h(E(\sigma_1)) \\
& = \sum^{|E_2|}_{j=1} h(E_1 \cup E_2^{[1,j]}) - h(E_1 \cup E_2^{[1,j-1]}) = \sum_{j=1}^{|E_2|} h(E_2^{[j,j]} | E_1 \cup E_2^{[1,j-1]}) \\
\end{aligned}
\]

There must exist an index $1\le j' \le |E_2|$ such that
\[
h(E_2^{[j',j']} | E_1 \oplus E_2^{[1,j'-1]}) \ge \frac{h(E(\sigma_2 \oplus \sigma_1)) - h(E(\sigma_1))}{|E_2|}
\]

Due to the submodularity of $h$, we have
\[
h(E_2^{[j',j']} | E_1) \ge h(E_2^{[j',j']} | E_1 \oplus E_2^{[1,j'-1]}) \ge \frac{1}{|E_2|}(h(E(\sigma_2 \oplus \sigma_1)) - h(E(\sigma_1)))
\]

Thus there exists some element $e$ and some vertex $v$ such that
\[
h(e|E_1) \ge \frac{1}{|E_2|}f(\sigma_2|\sigma_1) \Longrightarrow
f(v|\sigma_1) \ge \frac{1}{d_{\text{in}}|\sigma_2|}f(\sigma_2|\sigma_1)
\]
\end{proof}

\subsection{Proof of Lemma 2}
\textbf{Lemma 2.} \textit{
Consider $c \in (0,1]$ and $1 \le k' \le k$. Suppose that the sequence selected by the Sequence Greedy Algorithm is $\sigma$ with $|\sigma| = k$ and that there exists a sequence $\sigma'$ with $|\sigma'| = k-k'$ such that $\sigma' \subseteq \sigma$ and $f(\sigma') \ge c f(\sigma)$. Then we have $f(\sigma) \ge \frac{e^{\frac{k'}{d_{\text{in}}k}}-1} {e^{\frac{k'}{d_{\text{in}}k}}-c} f(\sigma^*(V,k,0))$.}
\begin{proof}
Let $v_2^i$ be the $i$-th vertex of sequence $\sigma_2$. Let $\sigma_2^i$ be the sequence consisting of the first $i$ vertices of sequence $\sigma_2$. We can safely assume that $|\sigma^*(V,k,0)|=k$.

\[
\begin{aligned}
f(\sigma_1 \oplus \sigma_2^i) - f(\sigma_1 \oplus \sigma_2^{i-1}) & = f(v_2^i|\sigma_1 \oplus \sigma_2^{i-1}) \ge h(e_2^i|E(\sigma_1 \oplus \sigma_2^{i-1})) \\
& \ge \frac{1}{d_{\text{in}}k} (h(E(\sigma^*(V,k,0)) \cup E(\sigma_1 \oplus \sigma_2^{i-1})) -h(E(\sigma_1 \oplus \sigma_2^{i-1})) \\
& = \frac{1}{d_{\text{in}}k} f(\sigma^*(V,k,0)|\sigma_1 \oplus \sigma_2^{i-1}) \ge \frac{1}{d_{\text{in}}k} (f(\sigma^*(V,k,0) - f(\sigma_1 \oplus \sigma_2^{i-1}))
\end{aligned}
\]
where $e_2^i$ is the edge that selected by the algorithm making $v_2^i$ added into the returned sequence. The above inequalities hold due to the fact that the function $f(\cdot)$ is monotone, Sequence Greedy algorithm select the edge of largest marginal value each time, and adding $\sigma^*(V,k,0)$ to $\sigma_1 \oplus \sigma_2^{i-1}$ may add at most $d_{\text{in}}k$ edges.

Rewriting the above inequality, we have
\[
f(\sigma_1 \oplus \sigma_2^i) \ge \frac{1}{d_{\text{in}}k} f(\sigma^*(V,k,0) + (1-\frac{1}{d_{\text{in}}k}) f(\sigma_1 \oplus \sigma_2^{i-1})
\]

Writing for $i\in \{1,2,...,k'\}$, we have
\[
\begin{aligned}
f(\sigma_1 \oplus \sigma_2^{k'}) &\ge \sum_{j=0}^{k'-1}\frac{1}{d_{\text{in}}k}(1-\frac{1}{d_{\text{in}}k})^j f(\sigma^*(V,k,0) + (1-\frac{1}{d_{\text{in}}k})^{k'} f(\sigma_1) \\
=& (1-(1- \frac{1}{d_{\text{in}}k})^{k'})f(\sigma^*(V,k,0) + (1-\frac{1}{d_{\text{in}}k})^{k'} f(\sigma_1)
\end{aligned}
\]
Thus we have
\[
\begin{aligned}
f(\sigma_2 | \sigma_1) = f(\sigma_1 \oplus \sigma_2^{k'}) - f(\sigma_1) & \ge (1-(1- \frac{1}{d_{\text{in}}k})^{k'})(f(\sigma^*(V,k,0)) - f(\sigma_1)) \\
& \ge (1-1/e^{\frac{k'}{d_{\text{in}}k}})(f(\sigma^*(V,k,0)) - f(\sigma_1))
\end{aligned}
\]

Suppose $f(\sigma)=\delta f(\sigma^*(V,k,0))$ for some $\delta\in(0,1]$. Then we have
\[
\begin{aligned}
\delta f(\sigma^*(V,k,0) & = f(\sigma) = f(\sigma_1)+f(\sigma_2|\sigma_1) \ge f(\sigma_1) + (1-1/e^{\frac{k'}{d_{\text{in}}k}})(f(\sigma^*(V,k,0)) - f(\sigma_1)) \\
& = (1/e^{\frac{k'}{d_{\text{in}}k}}) f(\sigma_1) + (1-1/e^{\frac{k'}{d_{\text{in}}k}})f(\sigma^*(V,k,0)) \ge \frac{c\delta}{e^{\frac{k'}{d_{\text{in}}k}}} f(\sigma^*(V,k,0)) + (1-\frac{1}{e^{\frac{k'}{d_{\text{in}}k}}})f(\sigma^*(V,k,0)) \\
\end{aligned}
\]

Then we have
\[
\delta \ge (c\cdot \delta/e^{\frac{k'}{d_{\text{in}}k}}) + (1-1/e^{\frac{k'}{d_{\text{in}}k}}) \Longrightarrow
\delta \ge \frac{e^{\frac{k'}{d_{\text{in}}k}}-1}{e^{\frac{k'}{d_{\text{in}}k}}-c}
\]
\end{proof}

\subsection{Proof of Lemma 3}
\textbf{Lemma 3.} \textit{
Consider $1 \le \tau \le k$. The following holds for any $Z \subseteq V$ with $|Z| \le \tau$: $g_\tau (\sigma^*(V,k,\tau)) \le f(\sigma^*(V-Z,k-\tau,0))$.}
\begin{proof}
Let $Z_{\tau}(\sigma)$ be the set that minimizes $g_{\tau}(\sigma)$. We can safely assume $|Z_{\tau}(\sigma)|=\tau$.

Let $\mathcal{U}=V(\sigma^*(V,k,\tau)) \cap Z$ and $\tau'=|\mathcal{U}|$. Obviously, $\mathcal{U}\subseteq V(\sigma^*(V,k,\tau))$. Also, we have
\[
\mathcal{U} \cup Z_{\tau-\tau'}(\sigma^*(V,k,\tau)-\mathcal{U}) \subseteq V(\sigma^*(V,k,\tau)) \quad\quad \text{and} \quad\quad |\mathcal{U} \cup Z_{\tau-\tau'}(\sigma^*(V,k,\tau)-\mathcal{U})| =\tau.
\]

Thus the set $\mathcal{U} \cup Z_{\tau-\tau'}(\sigma^*(V,k,\tau)-\mathcal{U})$ is a feasible solution to minimize function $g_\tau(\sigma)$ with $\sigma^*$ and $\tau$. Thus we have
\[
f(\sigma^*(V,k,\tau) -Z_\tau(\sigma^*(V,k,\tau))) \le f(\sigma^*(V,k,\tau) - \mathcal{U} \cup Z_{\tau-\tau'}(\sigma^*(V,k,\tau)-\mathcal{U}))
\]

In addition, we have $\sigma^*(V,k,\tau)-\mathcal{U} = \sigma^*(V,k,\tau) - Z$. This implies that
\[
Z_{\tau-\tau'}(\sigma^*(V,k,\tau) - \mathcal{U}) = Z_{\tau-\tau'}(\sigma^*(V,k,\tau) - Z).
\]

The elements in $Z-\mathcal{U}$ are not in $\sigma^*(V,k,\tau)$. Thus we have
\[
f( \sigma^*(V,k,\tau) - \mathcal{U} \cup Z_{\tau-\tau'}(\sigma^*(V,k,\tau) - \mathcal{U}) ) = f( \sigma^*(V,k,\tau) - Z \cup Z_{\tau-\tau'}(\sigma^*(V,k,\tau) - Z)
\]

Note the sequence $\sigma^*(V,k,\tau) - Z$ does not contain any element in $Z$ and has $k-\tau'$ elements. It is a feasible solution to minimize function $g_\tau(\sigma)$ with $V-Z,k-\tau'$ and $\tau-\tau'$. Thus we have
\[
g_{\tau-\tau'}(\sigma^*(V,k,\tau) - Z)  \le g_{\tau-\tau'}(\sigma^*(V-Z,k-\tau',\tau-\tau')) \le f(\sigma^*(V-Z,k-\tau,0))
\]

Also we have
\[
\begin{aligned}
g_{\tau-\tau'}(\sigma^*(V,k,\tau) - Z) & = f((\sigma^*(V,k,\tau) - Z) - Z_{\tau-\tau'}(\sigma^*(V,k,\tau) - Z)) \\
& = f((\sigma^*(V,k,\tau) - Z \cup Z_{\tau-\tau'}(\sigma^*(V,k,\tau) - Z)) \\
& = f((\sigma^*(V,k,\tau) - \mathcal{U} \cup Z_{\tau-\tau'}(\sigma^*(V,k,\tau) - \mathcal{U})) \\
& \ge f((\sigma^*(V,k,\tau) - Z_{\tau}(\sigma^*(V,k,\tau))) = g_{\tau}(\sigma^*(V,k,\tau))
\end{aligned}
\]
Hence we proved $g_{\tau}(\sigma^*(V,k,\tau)) \le f(\sigma^*(V-Z,k-\tau,0))$.
\end{proof}

\subsection{Proof of Theorem 2}
\textbf{Theorem 2.} \textit{
Consider $1\le \tau \le k$, Algorithm 1 achieves an approximation ratio of \[\max\{\frac{1-e^{-(1-1/k)}}{\alpha\beta},\frac{\tau\alpha\beta(\eta^{\frac{1}{d_{\text{in}}}}-1)}{\tau\alpha\eta^{\frac{1}{d_{\text{in}}}}- \beta (1-e^{-(1-1/k)}) }\}.\]}
\begin{proof}
After the execution of the RoseNets Algorithm, the selected sequence $\sigma=\sigma_1 \oplus \sigma_2$ where $|\sigma_1|=\tau$ and $|\sigma_2|=k-\tau$. Define $Z_{\tau} = Z_{\tau}^1 \cup Z_{\tau}^2$ as the removed set of vertices that can minimize $f(\sigma-Z_{\tau})$, where $Z_{\tau}^1 = Z_{\tau} \cap \sigma_1$ and $Z_{\tau}^2 = Z_{\tau} \cap \sigma_2$. First, we have a lower bound for $f(\sigma_2)$:
\[
f(\sigma_2) \ge \frac{1-e^{-(1-1/k)}}{\alpha} f(\sigma^*(V\backslash \sigma_1,k-\tau,0)) \ge \frac{1-e^{-(1-1/k)}}{\alpha} g_\tau(\sigma^*(V,k,\tau))
\]

The proof of Theorem 2 follows a similar line of the proof of Theorem 1. Specifically, we consider two cases as follows:

\textbf{Case 1. } $Z^2_{\tau}=\emptyset$, which means that all the removed vertices are in $\sigma_1$. Thus we have
\[
f(\sigma-Z_\tau(\sigma))=f(\sigma_2) \ge \frac{1-e^{-(1-1/k)}}{\alpha} g_\tau(\sigma^*(V,k,\tau))
\]

\textbf{Case 2. } $Z^2_{\tau} \neq \emptyset$. In this case, we further consider two cases.

\textbf{Case 2.1. } Let $f(\sigma_2) \le f(\sigma_2 - Z^2_\tau(\sigma))$.

In this case, the removal of elements in $Z^2_\tau(\sigma)$ does not reduce the overall value of the remaining sequence. Then we have
\[
\begin{aligned}
f(\sigma-Z_\tau(\sigma)) & = f((\sigma_1 - Z_\tau^1(\sigma)) \oplus (\sigma_2 - Z_\tau^2(\sigma)) \\
& \ge f(\sigma_2 -Z_\tau^2(\sigma)) \ge f(\sigma_2) \ge \frac{1-e^{-(1-1/k)}}{\alpha} g_\tau(\sigma^*(V,k,\tau))
\end{aligned}
\]

\textbf{Case 2.2. } Let $f(\sigma_2) \ge f(\sigma_2 - Z^2_\tau(\sigma))$. Let $\tau_1 = |Z_\tau^1(\sigma)|$ and $\tau_2=|Z_\tau^2(\sigma)|$. Then we have $\tau=\tau_1+\tau_2$ and $k=|\sigma|=|\sigma_1|+|\sigma_2|\ge \tau+\tau_2$.
%
%

Let $q= \frac{f(\sigma_2)-f(\sigma_2-Z_\tau^2(\sigma))}{(d_{\text{in}}+d_{\text{out}})f(\sigma_2)}$, denote the ratio of the loss of removing elements in $Z_\tau^2(\sigma)$ from sequence $\sigma_2$ to the value of the sequence $\sigma_2$. We have $q\in (0,\frac{1}{(d_{\text{in}}+d_{\text{out}})}]$ since $f(\sigma_2) > f(\sigma_2-Z_\tau^2(\sigma))$.

Let the elements in $Z^2_\tau(\sigma)$ be denoted by $z_1,z_2,...,z_{\tau_2}$ according to their order in sequence $\sigma_2$. Then we can rewrite $\sigma_2$ as $\sigma_2=\sigma_2^1\oplus z_1 \oplus \sigma_2^2 \oplus z_2 \oplus ... \oplus \sigma_2^{\tau_2+1}$, where $\sigma_2^i$ is the subsequence between item $z_{i-1}$ and $z_i$, for $i=1,2,...,\tau_2+1$ and $z_0$ and $z_{\tau_2+1}$ are empty sequences. Note $\sigma_2^i$ can be an empty sequence for any $i$. Then we have

\[
\begin{aligned}
(d_{\text{in}} & + d_{\text{out}}) q f(\sigma_2) = f(\sigma_2) - f(\sigma_2-Z_\tau^2(\sigma)) \\
& = f(\sigma_2^1) + f(z_1|\sigma_2^1)+f(\sigma_2^2| (\sigma_2^1\oplus z_1) + ... + f(\sigma_2^{\tau_2+1}| \sigma_2^1\oplus z_1 \oplus ... \oplus \sigma_2^{\tau_2})\\
& \quad - f(\sigma_2^1) -f(\sigma_2^2|\sigma_2^1) - ... - f(\sigma_2^{\tau_2+1}| \sigma_2^1\oplus ... \oplus \sigma_2^{\tau_2}) \\
& = \sum_{i=1}^{\tau_2} \left( f(z_i|\sigma_2^1\oplus z_1 \oplus ... \oplus z_{i-1}\oplus \sigma_2^i)+f(\sigma_2^{i+1}|\sigma_2^i\oplus z_i) - f(\sigma_2^{i+1}|\sigma_2^i) \right)\\
& \le \tau_2 d_{\text{in}} h(e_z^{\text{in}}) + \tau_2 d_{\text{out}} h(e_z^{\text{out}}) \\
& \le \tau_2(d_{\text{in}} + d_{\text{out}}) \max \{h(e_z^{\text{in}}),h(e_z^{\text{out}})\} \\
& \le \tau_2(d_{\text{in}} + d_{\text{out}}) f(Z_\tau^2)
\end{aligned}
\]
where $d_{\text{in}}$/$d_{\text{out}}$ is the maximum in/out degree of the network, $h(e_z^{\text{in}})$/$h(e_z^{\text{out}})$ are the edge that has maximum utility over all the incoming/outgoing edges of all the removed nodes $\{z_i\}_{i=1,2,...,\tau_2}$.  The first inequality is due to the fact that the marginal gain of a vertex $z$ to the prefix and subsequent sequence is at most $d_{\text{in}} h(e_z^{\text{in}})$ and $d_{\text{out}} h(e_z^{\text{out}})$. Then the second inequality follows intuitively.

Given Equation (4), we need to prove four inequalities for finally proving the theorem.

First, suppose the first vertex of $\sigma_2$ is $v_2$. By the monotonicity of function $f(\cdot)$ and the above inequality, we have
\[
\begin{aligned}
f(\sigma- Z_\tau(\sigma)) & = f((\sigma_1 - Z_\tau^1(\sigma)) \oplus (\sigma_2 - Z_\tau^2(\sigma)) \ge f(\sigma_1 - Z_\tau^1(\sigma)) = f(\sigma_1^{\tau_2}) \ge \tau_2 \max \{h(e_z^{\text{in}}),h(e_z^{\text{out}})\} \ge q f(\sigma_2) \\
f(\sigma-Z_\tau(\sigma)) & = f((\sigma_1 - Z_\tau^1(\sigma)) \oplus (\sigma_2 - Z_\tau^2(\sigma)) \ge f(\sigma_2 -Z_\tau^2(\sigma)) \ge (1-(d_{\text{in}} + d_{\text{out}})q) f(\sigma_2)
\end{aligned}
\]
Then we have Inequality 1 as below.

\textbf{Inequality 1:}
\[
f(\sigma-z)\ge \max\{ q \cdot f(\sigma_2),(1-(d_{\text{in}} + d_{\text{out}})q) \cdot f(\sigma_2)\}.
\]

By similar analysis in the proof of Theorem 1, we have inequality 2 as below.

\textbf{Inequality 2:}

\[
\max\{q,1-(d_{\text{in}} + d_{\text{out}})q\} \ge 1/\beta.
\]

Let $\sigma_2^{\tau_2}$ be the sequence consisting of the first $\tau_2$ elements of sequence $\sigma_2$. Then we have
\[
f(\sigma_2^{\tau_2}) \ge \frac{1-e^{-(1-1/k)}}{\alpha} f(\sigma^*(V\backslash \sigma_1,\tau_2,0)) \ge  \frac{1-e^{-(1-1/k)}}{\alpha} f(Z_\tau^2(\sigma)) \ge q \frac{1-e^{-(1-1/k)}}{\tau_2 \alpha} f(\sigma_2)
\]

Thus we have
\[
\begin{aligned}
f(\sigma_2) & \ge \frac{e^{\frac{k-\tau-\tau_2-1}{(k-\tau_2-1)d_{\text{in}}}}-1}{e^{\frac{k-\tau-\tau_2-1}{(k-\tau_2-1)d_{\text{in}}}}- q \frac{1-e^{-(1-1/k)}}{\tau_2\alpha} } f(\sigma^*(V\backslash \sigma_1,k-\tau,0)) \\
\Longrightarrow & \text{\textbf{ Inequality 3: }} f(\sigma_2) \ge \frac{e^{\frac{k-\tau-\tau_2-1}{(k-\tau_2-1)d_{\text{in}}}}-1}{e^{\frac{k-\tau-\tau_2-1}{(k-\tau_2-1)d_{\text{in}}}}- q \frac{1-e^{-(1-1/k)}}{\tau_2\alpha} } g_\tau(\sigma^*(V,k,\tau))
\end{aligned}
\]

Now define $\ell_1(q) = q \cdot \frac{e^{\frac{k-\tau-\tau_2-1}{(k-\tau_2-1)d_{\text{in}}}}-1}{e^{\frac{k-\tau-\tau_2-1}
{d_{\text{in}}(k-\tau_2-1)}}-q \frac{1-e^{-(1-1/k)}}{\tau_2\alpha} } $ and $\ell_2(q) = (1-(d_{\text{in}} + d_{\text{out}})q) \frac{e^{\frac{k-\tau-\tau_2-1}{(k-\tau_2-1)d_{\text{in}}}}-1}{e^{\frac{k-\tau-\tau_2-1}
{d_{\text{in}}(k-\tau_2-1)}}-q \frac{1-e^{-(1-1/k)}}{\tau_2\alpha} }$. It is easy to verify that for $k\ge 3$ and $q\in (0,\frac{1}{d_{\text{in}}+d_{\text{out}}}]$, $\ell_1(q)$/$\ell_2(q)$ is monotonically increasing/decreasing. Note when $q=\frac{1}{\beta}$, $\ell_1(q)=\ell_2(q)=\frac{\beta(e^{\frac{k-\tau-\tau_2-1}{(k-\tau_2-1)d_{\text{in}}}}-1)}{e^{\frac{k-\tau-\tau_2-1}
{d_{\text{in}}(k-\tau_2-1)}}-\beta \frac{1-e^{-(1-1/k)}}{\tau_2\alpha} }$. We consider two cases for $q$: (1) when $q \in (0,\frac{1}{\beta}]$, we have $\max\{\ell_1(q),\ell_2(q)\}\ge \ell_2(\frac{1}{\beta})$  as $\ell_2(q)$ is monotonically decreasing; (2) when $q \in (\frac{1}{\beta},\frac{1}{d_{\text{in}}+\text{out}}]$, we have $\max\{\ell_1(q),\ell_2(q)\}\ge \ell_1(\frac{1}{\beta})$ as $\ell_1(q)$ is monotonically increasing. Thus we have the Inequality 4 as below.

\textbf{Inequality 4:}
\[
\max\{ q \cdot \frac{e^{\frac{k-\tau-\tau_2-1}{(k-\tau_2-1)d_{\text{in}}}}-1}{e^{\frac{k-\tau-\tau_2-1}
{d_{\text{in}}(k-\tau_2-1)}}-q \frac{1-e^{-(1-1/k)}}{\tau_2\alpha} } , (1-(d_{\text{in}} + d_{\text{out}})q) \frac{e^{\frac{k-\tau-\tau_2-1}{(k-\tau_2-1)d_{\text{in}}}}-1}{e^{\frac{k-\tau-\tau_2-1}
{d_{\text{in}}(k-\tau_2-1)}}-q \frac{1-e^{-(1-1/k)}}{\tau_2\alpha} } \} \ge
\frac{\beta(e^{\frac{k-\tau-\tau_2-1}{(k-\tau_2-1)d_{\text{in}}}}-1)}{e^{\frac{k-\tau-\tau_2-1}
{d_{\text{in}}(k-\tau_2-1)}}-\beta \frac{1-e^{-(1-1/k)}}{\tau_2\alpha} }
\]

Combining Inequality 1, Inequality 2 and the lower bound of $f(\sigma_2)$, we can have the first lower bound in Theorem 2:
\[
\begin{aligned}
f(\sigma -z ) & \ge \max\{ q \cdot f(\sigma_2),(1-(d_{\text{in}} + d_{\text{out}})q) \cdot f(\sigma_2)\} \\
& \ge \max\{q,1-(d_{\text{in}} + d_{\text{out}})q\} \cdot \frac{1-e^{-(1-1/k)}}{\alpha} g_\tau(\sigma^*(V,k,\tau)) \\
& \ge \frac{1-e^{-(1-1/k)}}{\alpha\beta} g_\tau(\sigma^*(V,k,\tau))
\end{aligned}
\]

Combining Inequality 1, Inequality 3 and Equation Inequality 4, we can have the second lower bound in Theorem 2:
\[
\begin{aligned}
f(\sigma -z ) & \ge \max\{ q \cdot f(\sigma_2),(1-(d_{\text{in}} + d_{\text{out}})q) \cdot f(\sigma_2)\} \\
& \ge \max\{ q \cdot \frac{e^{\frac{k-\tau-\tau_2-1}{(k-\tau_2-1)d_{\text{in}}}}-1}{e^{\frac{k-\tau-\tau_2-1}{(k-\tau_2-1)d_{\text{in}}}} q \frac{1-e^{-(1-1/k)}}{\alpha} } , (1-(d_{\text{in}} + d_{\text{out}})q) \frac{e^{\frac{k-\tau-\tau_2-1}{(k-\tau_2-1)d_{\text{in}}}}-1}{e^{\frac{k-\tau-\tau_2-1}{(k-\tau_2-1)d_{\text{in}}}}-\tau_2 q \frac{1-e^{-(1-1/k)}}{\alpha} } \} g_\tau(\sigma^*(V,k,\tau)) \\
& \ge \frac{\beta(e^{\frac{k-\tau-\tau_2-1}{(k-\tau_2-1)d_{\text{in}}}}-1)}{e^{\frac{k-\tau-\tau_2-1}{(k-\tau_2-1)d_{\text{in}}}}- \beta \frac{1-e^{-(1-1/k)}}{\tau_2\alpha} } g_\tau(\sigma^*(V,k,\tau))
\end{aligned}
\]
Since $\tau_2 \le \tau$, the theorem holds.
\end{proof}

\end{document}